\documentclass{article} 
\PassOptionsToPackage{table}{xcolor}
\usepackage{iclr2027_conference,times}

\usepackage{amsmath,amsfonts,bm}

\def\eqref#1{equation~\ref{#1}}

\def\1{\bm{1}}

\DeclareMathAlphabet{\mathsfit}{\encodingdefault}{\sfdefault}{m}{sl}
\SetMathAlphabet{\mathsfit}{bold}{\encodingdefault}{\sfdefault}{bx}{n}

\usepackage[utf8]{inputenc}
\usepackage[T1]{fontenc}
\usepackage{hyperref}
\usepackage{url}
\usepackage{booktabs}
\usepackage{amsfonts}
\usepackage{amsmath}
\usepackage{microtype}
\usepackage{xcolor}
\usepackage{array}
\usepackage{graphicx}
\usepackage{wrapfig}
\usepackage{makecell}
\usepackage{verbatim}
\usepackage{listings}
\usepackage[breakable,skins]{tcolorbox}

\newcommand{\mypara}[1]{\vspace{0.3mm}\noindent\textbf{#1}\hskip1em}

\def\dataset{{ChartGalaxy++}}

\newcommand{\totalnum}{217,195}
\newcommand{\realnum}{56,744}
\newcommand{\syntheticnum}{160,451}

\newcommand{\elementnum}{15,591,656}
\newcommand{\textelementnum}{6,044,031}
\newcommand{\imageelementnum}{1,107,791}
\newcommand{\shapeelementnum}{8,439,834}
\newcommand{\groupnum}{3,305,827}
\newcommand{\hierarchicalrelationnum}{18,680,288}
\newcommand{\spatialrelationnum}{21,309,060}
\newcommand{\avgelementnum}{71.8}
\newcommand{\avggroupnum}{15.2}
\newcommand{\avggraphdepth}{4.5}

\title{\dataset{}: A Richly Annotated Dataset for Chart Understanding and Generation}

\author{Duan Li$^{1}$* \quad Zheng Wang$^{1}$* \quad Jiangning Zhu$^{1}$* \quad Yuhui Yuan$^{2}$ \quad Shixia Liu$^{1}$\textdagger \\
$^{1}$Tsinghua University \quad $^{2}$Canva \\
* Equal contribution. \\
\textdagger\ Corresponding author.}

\iclrfinalcopy 

\begin{document}

\maketitle
\fancyhead[L]{} 

\begin{figure}[ht]
\centering
\setlength{\abovecaptionskip}{1mm}
\includegraphics[width=\linewidth]{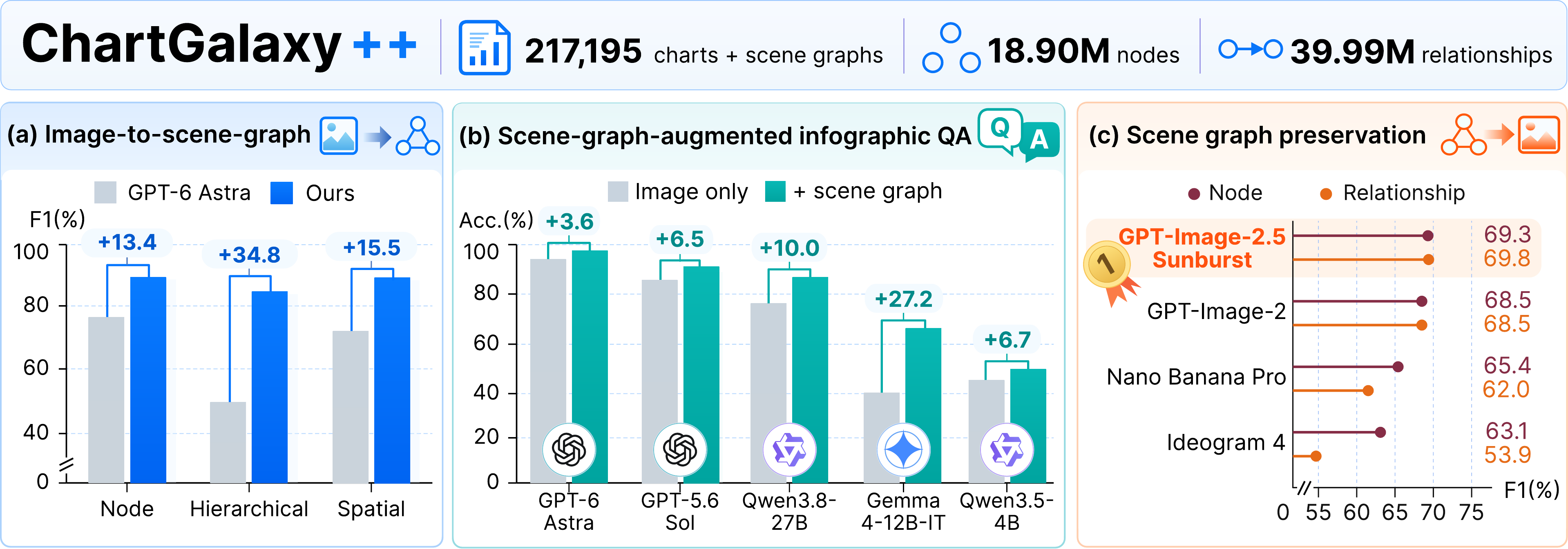}
\caption{\dataset{} provides scene graph annotations for \totalnum{} infographic charts. 
These annotations support (a) image-to-scene-graph prediction, (b) scene-graph-augmented infographic question answering, and (c) evaluation of scene graph preservation in image generation.
}
\label{fig:abs}
\end{figure}

\begin{abstract}
Infographic charts communicate data through cohesive compositions of visual elements such as charts, text, and images.
Existing infographic chart datasets primarily annotate individual elements with bounding boxes and category labels.
Such element-level annotations do not capture the hierarchical and spatial relationships, limiting their use for understanding how elements are organized and related within an infographic and generating well-organized designs.
To address this limitation, we introduce \dataset{}, a richly annotated dataset of \totalnum{} infographic charts.
At the core of this dataset is a scene graph representation, in which nodes represent visual elements and semantic groups, and edges encode their hierarchical and spatial relationships.
For real infographics, we construct scene graphs by detecting and progressively grouping visual elements; for synthetic ones, we derive them directly from the generation process.
We demonstrate the utility of \dataset{} through three applications: 1) image-to-scene-graph prediction, 2) scene-graph-augmented infographic question answering, and 3) evaluation of scene graph preservation in image generation.
Together, these applications highlight the value of \dataset{} for advancing chart understanding and generation.

\raisebox{-0.3\height}{\hspace{0.07cm}\includegraphics[width=0.40cm]{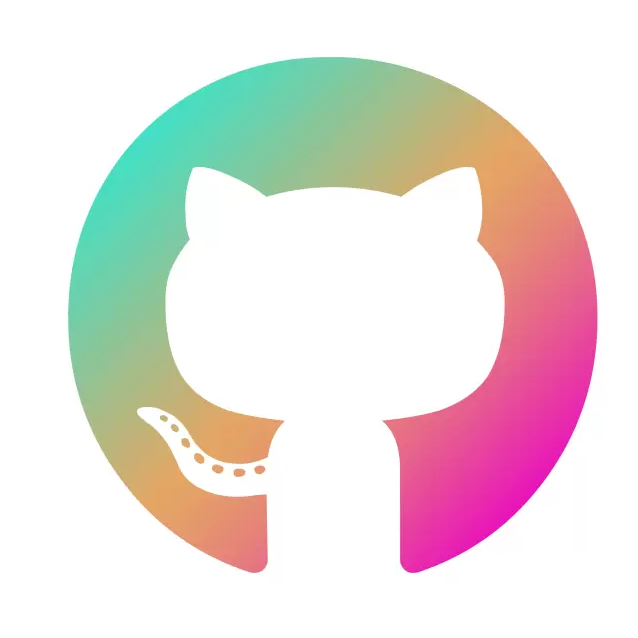}} \small \textbf{\mbox{Code:}} \href{https://github.com/ChartGalaxyPP/ChartGalaxyPlusPlus}{https://github.com/ChartGalaxyPP/ChartGalaxyPlusPlus} \\
\vspace{1em}
\raisebox{-0.3\height}{\includegraphics[width=0.42cm]{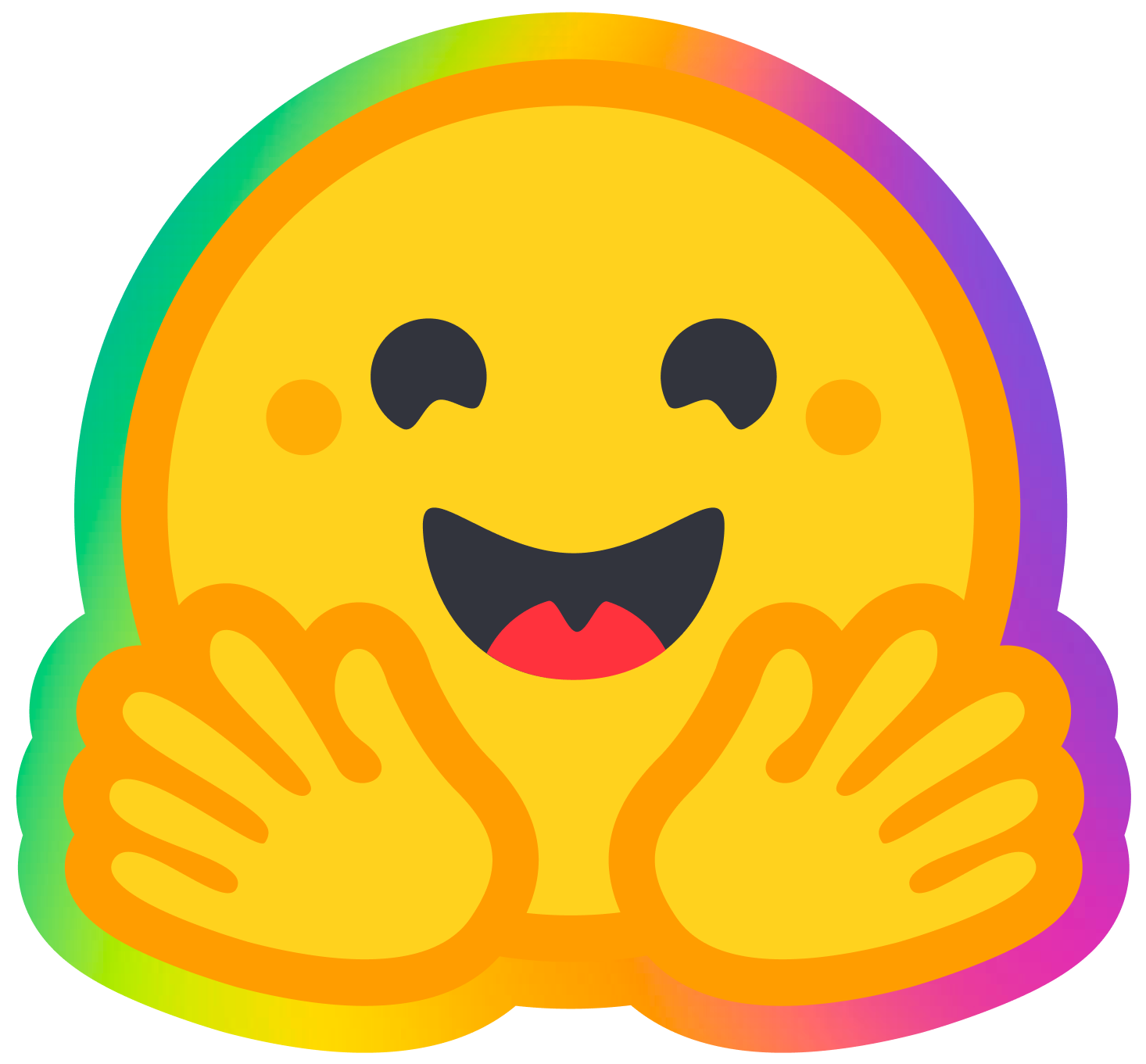}} \small \textbf{\mbox{Data \& Dataset Card:}} \href{https://huggingface.co/datasets/ChartGalaxyPP/ChartGalaxyPlusPlus}{https://huggingface.co/datasets/ChartGalaxyPP/ChartGalaxyPlusPlus}
\end{abstract}

\section{Introduction}
\label{sec:introduction}

Infographic charts integrate visual elements, such as charts, text, and images, into cohesive visual designs.
By presenting complex data in an accessible and compelling manner, they are widely used in domains such as news reporting, business communication, and public education~\citep{cui2022mixed,elaldi2021effectiveness}.
Despite these benefits, the visual richness and structural complexity of infographic charts pose substantial challenges for foundation models in both understanding and generation.
For understanding, even frontier Large Vision-Language Models (LVLMs), such as GPT-5.6~\citep{openai2026gpt56}, still struggle to locate relevant evidence in complex charts, integrate information across elements, and perform reliable multi-step reasoning over the underlying data~\citep{masry2025chartqapro,efat2026polychartqa,xiao2026longchart}.
For generation, frontier image-generation models, such as Nano Banana Pro~\citep{google2025nanobananapro} and GPT-Image-2~\citep{openai2026gptimage2}, can produce visually appealing outputs but still struggle to faithfully represent the underlying data, accurately render visual elements, and organize them into coherent designs~\citep{tang2026igenbench}.
These limitations suggest that understanding and generating infographic charts require models to capture not only individual elements but also the relationships that organize them into a coherent whole.

Recent studies have made progress toward this goal by constructing large-scale infographic datasets with element-level annotations~\citep{zhu2026infodet,gutflaish2025structuredcaptions,li2026chartgalaxy}.
For example, \cite{zhu2026infodet} effectively improve the understanding performance of LVLMs by providing bounding-box annotations for charts.
Despite this progress, existing datasets primarily annotate visual elements independently and do not explicitly model the relationships among them.
These relationships can be either hierarchical, specifying how individual elements are organized into groups, or spatial, characterizing their relative direction, alignment, overlap, and proximity~\citep{shi2023reverse}.
Both provide structural information beyond element-level annotations.
Such structural information can, for example, help associate a value label with the correct data mark and preserve their alignment during generation.

To explicitly model these relationships, we introduce a scene graph representation in which leaf nodes represent individual visual elements, internal nodes represent semantic groups, and edges encode hierarchical and spatial relationships.
Based on this representation, we construct \dataset{} by augmenting the real and synthetic infographic charts in ChartGalaxy~\citep{li2026chartgalaxy} with scene graph annotations.
For real infographic charts, we first extract text and non-text elements using HunyuanOCR-1.5~\citep{HunyuanOCR_1_5_2026} and DAD~\citep{zhu2026dad}, respectively.
We then use Gemini 3.8 Flash~\citep{google2026gemini38flash} to annotate element attributes and extract hierarchical relationships by organizing individual elements into semantic groups.
Spatial relationships among elements and groups are subsequently derived from their bounding boxes.
The resulting annotations are iteratively refined based on errors identified through programmatic checks and LVLM-based verification.
For synthetic infographic charts,  scene graphs are directly extracted from the generation process, where the underlying element structure and layout information are readily available.
In total, \dataset{} contains \totalnum{} infographic charts, including \realnum{} real ones and \syntheticnum{} synthetic ones.
Their scene graphs comprise 18.90 million nodes and 39.99 million edges.

We evaluate the utility of \dataset{} through three applications (Fig.~\ref{fig:abs}).
First, we develop an image-to-scene-graph model to predict visual elements and their relationships.
Second, we augment infographic question answering (QA) with scene graphs to assess whether explicit structural information improves evidence grounding and reasoning.
Third, we evaluate whether frontier image-generation models can preserve the elements and relationships specified by an input scene graph.
Together, these applications demonstrate the value of \dataset{} for developing and evaluating models that extract, use, and preserve infographic structure for understanding and generation.

The main contributions of our work include:

\setlength{\leftmargini}{2.5em}
\begin{itemize}
\item A scene graph representation for infographic charts that organizes visual elements into semantic groups and models their hierarchical and spatial relationships.
\item An automated pipeline for constructing scene graph annotations for both real and synthetic infographic charts.
\item A large-scale dataset of \totalnum{} infographic charts with rich scene graph annotations for structure-aware chart understanding and generation.

\end{itemize}

\section{Related Work}
\label{sec:related-work}

A growing number of datasets provide element-level annotations for plain and infographic charts.

Plain charts present data in a minimal manner using geometric shapes and text.
Earlier plain-chart datasets mainly provide bounding boxes and category annotations for these visual elements~\citep{kahou2017figureqa,methani2020plotqa,deng2022visimages,dou2024hierarchically}.
For example, PlotQA~\citep{methani2020plotqa} contains 224,377 programmatically generated charts with the bounding boxes and categories extracted from the generation process.
To support the understanding of chart semantics, more recent datasets introduce grounding annotations that identify elements relevant to a given query~\citep{vogel2025refchartqa,suri2025chartlens,bansal2025chartab,ito2026phrase,liu2026start,niu2026chartreg} or reasoning annotations that associate intermediate reasoning steps with the involved elements~\citep{xu2025chartpoint,lee2026visdot}.
Beyond element-level annotations, several datasets also capture chart structure.
VisText~\citep{tang2023vistext} consists of 12,441 charts programmatically generated using Vega-Lite and provides hierarchical representations based on parent-child relationships.
VisAnatomy~\citep{chen2026visanatomy} consists of 942 real-world SVG charts created with over 50 visualization tools, with manually annotated hierarchical element groupings and group layouts.
While these datasets provide rich structural annotations for plain charts, they do not capture more diverse visual elements and relationships in infographic charts.

To bridge this gap, several datasets annotate elements in infographic charts~\citep{borkin2015beyond,li2026chartgalaxy,zhu2026infodet}.
ChartGalaxy~\citep{li2026chartgalaxy} contains over 1.7 million infographic charts collected online or generated from layout templates, with bounding-box annotations for text, images, and charts.
InfoDet~\citep{zhu2026infodet} further provides finer-grained annotations for synthetic infographics by additionally labeling 26 categories of chart marks, such as bars, axes, and legends.
However, these datasets still treat elements largely in isolation and do not explicitly model how they are grouped and spatially organized.
\dataset{} addresses this limitation by introducing scene-graph-based rich annotations that capture hierarchical and spatial relationships among elements and groups, enabling structure-aware infographic understanding and generation.

\section{Scene Graph Representation}
\label{sec:scene_graph}

\begin{figure}[!t]
\centering
\includegraphics[width=\linewidth]{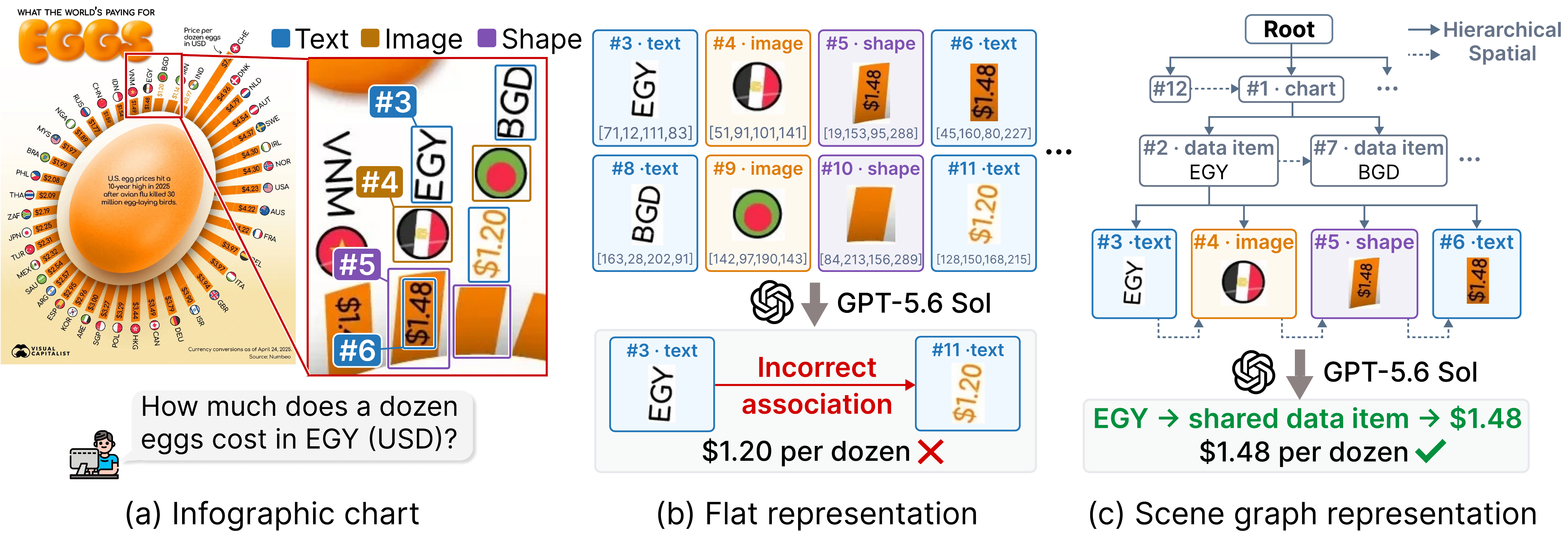}
\caption{Comparison of flat and scene graph representations for infographic understanding.
Flat representations lack explicit structural information, potentially leading to incorrect element associations. 
Our scene graph representation resolves such ambiguities by explicitly grouping semantically related elements and encoding their hierarchical and spatial relationships.
}
\label{fig:scene_graph_representation}
\end{figure}

Existing infographic-chart representations typically model an image as a flat collection of categorized bounding boxes. 
While such representations capture individual elements and their locations, they do not encode an important characteristic of infographics: visual elements are compositionally organized into semantic groups, and these elements and groups are spatially organized into coherent layouts. 
Without explicitly modeling this structure, models must infer element associations and layout organization from isolated elements, making it difficult to distinguish visually adjacent but semantically unrelated elements.
For example, as shown in Fig.~\ref{fig:scene_graph_representation}(b), the flat representation provides no explicit indication of which price is associated with EGY. 
Consequently, GPT-5.6 sol may associate EGY with the spatially nearby price of BGD (\$1.20), rather than its corresponding price (\$1.48).

To address this issue, we introduce a scene graph representation that 
transforms a flat collection of visual elements into a structured composition of semantic groups. 
Specifically, it groups elements that jointly encode a data item into semantic groups, recursively organizes these groups into a hierarchy, and encodes spatial relationships among elements and groups.
Formally, we represent each infographic chart as a scene graph $\mathcal{G}=(V,E)$, where $V$ and $E$ denote the node and edge sets, respectively.
The node set $V=V_{\ell}\cup V_i$ contains leaf and internal nodes.
Each leaf node $v\in V_{\ell}$ corresponds to a visual element, such as a text block, image, or geometric shape.
Each internal node $v\in V_i$ represents a group containing multiple elements and, recursively, other groups.
For example, in Fig.~\ref{fig:scene_graph_representation}(c), the country label, flag, bar mark, and value label associated with Egypt are grouped under the same internal node because they jointly encode one data item.
This explicit grouping distinguishes the value ``\$1.48'' associated with EGY from nearby values belonging to other countries.

Each node $v\in V$ stores four common attributes: type, bounding box, color palette, and description.
Text nodes additionally store their text content.
The edge set $E=E_h\cup E_s$ contains hierarchical and spatial-relationship edges.
Hierarchical edges $E_h$ encode parent-child relationships between nodes, forming a nested semantic hierarchy.
Spatial-relationship edges $E_s$ encode geometric relationships between node pairs, including relative direction, alignment, overlap, and proximity.
The detailed schema for nodes and edges is provided in Appendix~\ref{app:schema}.

\section{Dataset Construction}
\label{sec:dataset}

\dataset{} builds on ChartGalaxy~\citep{li2026chartgalaxy}, which contains both real infographic charts collected from the web and synthetic ones programmatically generated from layout templates.
We augment both types with scene graph annotations that jointly capture visual elements, as well as their hierarchical and spatial relationships.

\subsection{Annotation Pipeline}
\label{sec:annotation}

\begin{figure}[!t]
\centering
\includegraphics[width=\linewidth]{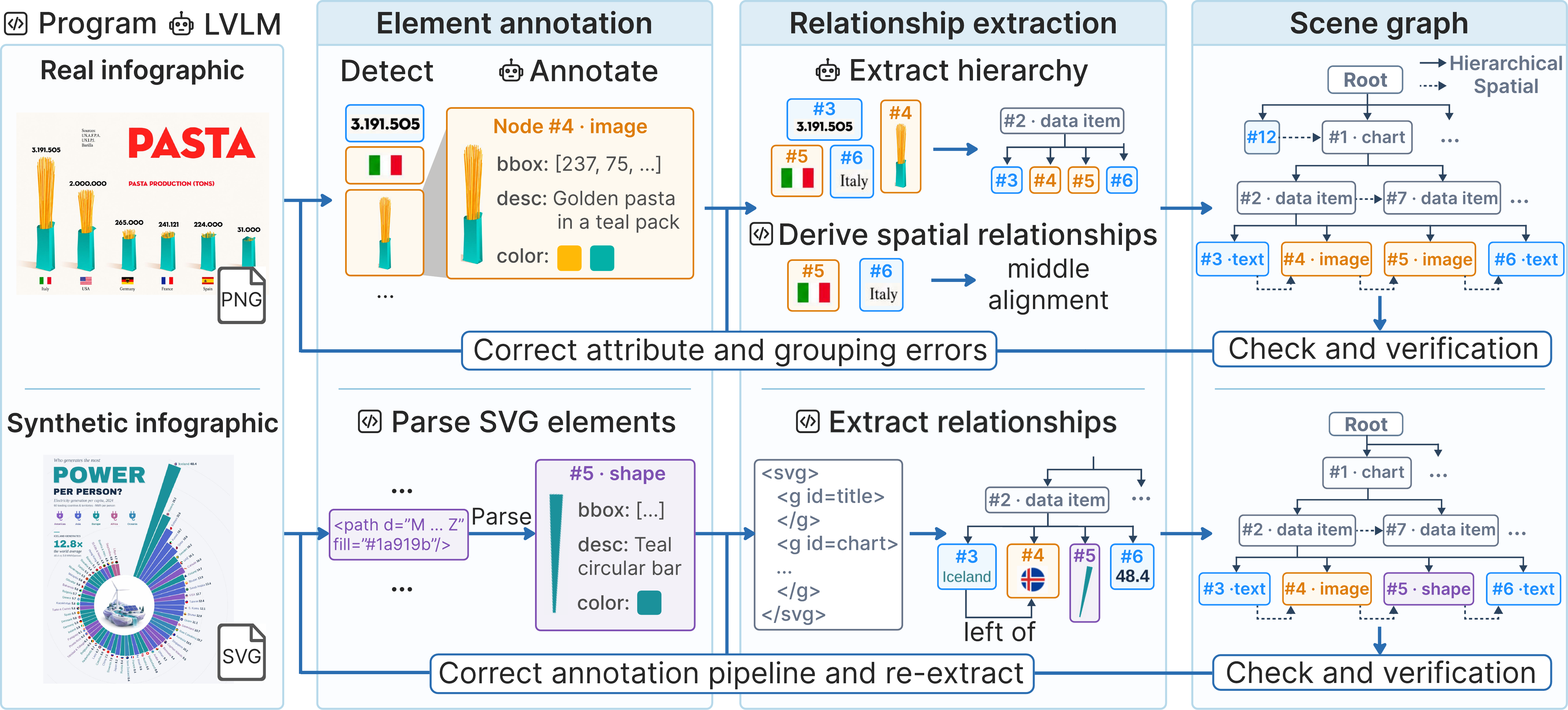}
\caption{Overview of the scene graph annotation pipeline.}
\label{fig:annotation_pipeline}
\end{figure}

Real infographic charts are available only as images, while synthetic ones retain their underlying element structure and layout information from the generation process.
We therefore use a dedicated annotation pipeline for each source (Fig.~\ref{fig:annotation_pipeline}).

\mypara{Real infographic chart annotation}
Given a real infographic chart, our scene graph annotation pipeline consists of two stages: element annotation and relationship extraction.
We then iteratively refine the resulting scene graph by identifying errors and correcting them in the corresponding stage.

\textit{Element annotation}.
We first extract the content and bounding boxes of text elements using HunyuanOCR-1.5~\citep{HunyuanOCR_1_5_2026}, and the bounding boxes of non-text elements using DAD~\citep{zhu2026dad}.
Given the image and extracted bounding boxes, we then use Gemini 3.8 Flash~\citep{google2026gemini38flash} to annotate each element's type, color palette, and description.
These annotations define the leaf nodes of the scene graph.

\textit{Relationship extraction}.
Extracting hierarchical and spatial relationships relies on different types of information.
Hierarchical relationships depend on visible region boundaries, relative placement, and how elements jointly convey information.
We therefore use Gemini 3.8 Flash to integrate such information and group elements into a hierarchy.
Each group is represented as an internal node connected to its children by hierarchical edges.
Spatial relationships can instead be determined directly from node bounding boxes, so we derive them programmatically.
Together, the annotated nodes and extracted relationships form the initial scene graph.

\textit{Iterative refinement}.
We validate the initial scene graph using programmatic checks and LVLM-based verification.
The programmatic check detects invalid attribute values, unresolved node references, and violations of hierarchy constraints.
The LVLM-based verification uses Gemini 3.8 Flash to compare the scene graph with the original infographic to identify missing elements, incorrect attributes, and incorrect groupings.
The detected errors are corrected in the corresponding annotation stages, and the updated scene graph is re-verified until no further errors are identified.

The implementation details for real infographic annotation are provided in Appendix~\ref{app:annotation}.

\mypara{Synthetic infographic chart annotation}
ChartGalaxy generates synthetic infographic charts as SVG files, which explicitly encode visual elements and their layout information.
We therefore extract the scene graph directly from these files.

\textit{Element annotation.}
We parse the SVG to extract text, images, and geometric shapes.
For each element, we compute its bounding box from its position, size, and transformations encoded in the SVG, and extract other attributes from the corresponding SVG tags and metadata.
These extracted elements form the leaf nodes of the scene graph.

\textit{Relationship extraction.}
We derive internal nodes and hierarchical edges from the grouping structure of the SVG. 
We then derive spatial relationships from the bounding boxes of the resulting nodes.

\textit{Iterative refinement}.
The annotated nodes and extracted relationships form the scene graph.
We validate the resulting scene graphs using the programmatic check and LVLM-based verification.
If either identifies a potential error, we manually inspect the corresponding infographic and update the annotation pipeline as needed.

\subsection{Data Statistics}
\label{sec:statistics}
\dataset{} comprises \totalnum{} infographic charts, including \realnum{} real and \syntheticnum{} synthetic ones, each paired with its scene graph annotation.
These scene graph annotations include \elementnum{} elements: \textelementnum{} text blocks, \imageelementnum{} images, and \shapeelementnum{} geometric shapes, averaging \avgelementnum{} elements per chart.
They further encode \hierarchicalrelationnum{} hierarchical and \spatialrelationnum{} spatial relationships.
On average, each scene graph consists of \avggroupnum{} groups and spans \avggraphdepth{} hierarchy levels.
The detailed statistics are provided in Appendix~\ref{app:dataset-details}.
For consistent evaluation, we use 1,000 infographic charts as the test set and the remainder as the training set.
The test set annotations are manually verified.



\section{Experiments}
\label{sec:experiments}


We evaluate the utility of \dataset{} through three applications: image-to-scene-graph prediction, scene-graph-augmented QA, and scene graph preservation in image generation.

\subsection{Image-to-Scene-Graph Prediction}
\label{subsec:i2jmodel}

\subsubsection{Experimental Setup}

\mypara{Training setup}
We build an image-to-scene-graph model by supervised fine-tuning Qwen3.5-4B~\citep{qwen2026qwen35} on the \dataset{} training set.
Since spatial relationships can be easily derived from bounding boxes using deterministic rules, we train the model to predict only the remaining scene graph annotations and compute spatial relationships programmatically from its predicted bounding boxes.
We perform full-parameter fine-tuning for $100,000$ steps with a batch size of 32 and a learning rate of $5e{-}5$.
Additional training details are provided in Appendix~\ref{app:i2sg-training}.

\mypara{Baseline models}
We compare our model with ten representative open-source and proprietary LVLMs.
Specifically, we select seven open-source models (Molmo2-8B~\citep{allenai2026molmo2}, InternVL3.5-8B~\citep{opengvlab2025internvl35}, Gemma~4-12B-IT~\citep{google2026gemma4}, Qwen3.5-4B~\citep{qwen2026qwen35}, Qwen3.8-27B~\citep{qwen2026qwen38dense}, Qwen3.8-Flash-Next-FP8~\citep{qwen2026qwen38flashnext}, and GLM-5.3-Flash~\citep{zai2026glm53flash}) and three proprietary ones (Gemini~3.8 Flash~\citep{google2026gemini38flash}, GPT-5.6 Sol~\citep{openai2026gpt56}, and GPT-6 Astra~\citep{openai2026gpt6astra}).
For each image, all models receive a prompt that specifies the scene graph schema and asks the model to predict the scene graph excluding spatial relationships.
We then derive the spatial relationships programmatically from the predicted bounding boxes.
Additional details of the baseline settings are provided in Appendix~\ref{app:i2sg-baselines}.

\mypara{Evaluation metrics}
We evaluate predicted scene graphs against their ground-truth annotations at two levels: nodes and edges.

\textit{Node-level metrics}.
For each pair of predicted and ground-truth nodes, we compute similarities for the following attributes:
1) type, measured by exact matching; 
2) bounding box, measured by generalized intersection over union (GIoU)~\citep{Rezatofighi2019GIoU}, which considers both overlap and spatial separation; 
3) text content, for text nodes, using normalized Levenshtein similarity~\citep{levenshtein1966binary}, which reflects character-level edit distance; 
4) color palette, measured by matching colors using CIEDE2000 color difference~\citep{luo2001development} and assessing whether predicted colors remain within and cover the ground-truth palette; and 
5) description, evaluated by GPT-5.6 Luna~\citep{openai2026gpt56luna} based on semantic similarity.
The similarity between two nodes is computed by averaging the applicable attribute similarities.
Implementation details of these metrics are provided in Appendix~\ref{app:metrics}.

Following \cite{jiang2026rexomni}, we evaluate overall node-level performance using a similarity-weighted F1 score.
We first use the Hungarian algorithm to find the one-to-one matching that maximizes the total similarity between predicted and ground-truth nodes.
We then compute precision as the summed similarities of matched pairs divided by the number of predicted nodes, and recall as the same sum divided by the number of ground-truth nodes.
Their harmonic mean is used as the F1 score, and the mean similarity of each attribute over the matched pairs is also reported.

\textit{Edge-level metrics}.
Following \cite{wu2021screen}, we evaluate edge-level performance using the F1 score of predicted edges.
A predicted edge is a true positive if both endpoint nodes are matched and the same relationship exists between their corresponding ground-truth nodes.
All other predicted edges are false positives, and ground-truth edges without a corresponding true positive are false negatives.
We report F1 scores separately for hierarchical and spatial edges.

\begin{table}[!t]
\centering
\caption{Comparison of our image-to-scene-graph model with representative LVLMs.}
\label{tab:image2sg}
\renewcommand{\arraystretch}{1.2}
\resizebox{\linewidth}{!}{%
    \begin{tabular}{l|c|ccccc|cc}
        \hline
        & \multicolumn{6}{c|}{Node-level}
        & \multicolumn{2}{c}{Edge-level} \\
        \hline
        Model
        & Overall
        & Type
        & Box
        & Text
        & Color
        & Description
        & Hierarchical
        & Spatial\\
        \hline
        Molmo2-8B & 8.7 & 67.3 & 32.0 & 43.2 & 65.0 & 34.9 & 4.0 & 0.9 \\
        InternVL3.5-8B & 27.7 & 79.6 & 34.6 & 75.1 & 68.7 & 36.2 & 9.8 & 10.0 \\
        Gemma~4-12B-IT & 45.8 & 94.1 & 51.1 & 91.8 & 77.5 & 53.7 & 18.1 & 29.5 \\
        Qwen3.5-4B & 27.8 & 90.5 & 25.3 & 86.4 & 80.1 & 51.3 & 9.0 & 10.6 \\
        Qwen3.8-27B & 60.0 & 95.6 & 65.7 & \underline{96.5} & 87.3 & 54.8 & 27.3 & 47.1 \\
        Qwen3.8-Flash-Next-FP8 & 60.9 & 95.6 & 68.8 & 96.2 & 86.3 & 54.3 & 32.3 & 50.2 \\
        GLM-5.3-Flash & 63.1 & 93.5 & 66.5 & 95.4 & 88.6 & 60.3 & 34.9 & 50.1 \\
        \hline
        Gemini~3.8 Flash & 70.1 & 95.2 & 83.8 & \underline{96.5} & 90.3 & \underline{64.1} & 47.7 & 61.5 \\
        GPT-5.6 Sol & 69.8 & 95.4 & 77.5 & 94.9 & 91.6 & 59.5 & 45.8 & 63.3 \\
        GPT-6 Astra & \underline{76.0} & \underline{95.7} & \underline{87.4} & 95.6 & \underline{91.9} & 59.0 & \underline{50.3} & \underline{72.5} \\
        \hline
        \rowcolor{blue!10} Ours & \textbf{89.4} & \textbf{98.5} & \textbf{95.4} & \textbf{97.2} & \textbf{96.6} & \textbf{90.2} & \textbf{85.1} & \textbf{88.0} \\
        \hline
    \end{tabular}%
}
\end{table}

\begin{figure}[!t]
\centering
\includegraphics[width=\linewidth]{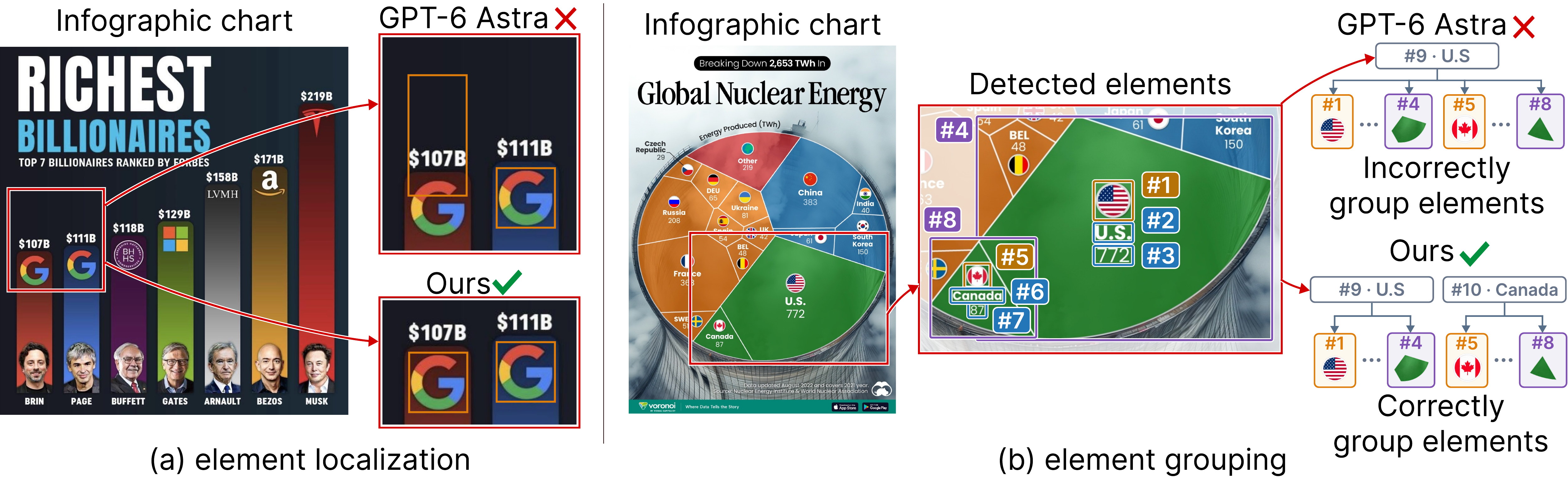}
\caption{Qualitative comparisons of our image-to-scene-graph model and GPT-6 Astra.}
\label{fig:qualitative}
\end{figure}

\subsubsection{Results and analysis}
Table~\ref{tab:image2sg} compares our image-to-scene-graph model with the baselines on the \dataset{} test set.
Our model achieves the highest scores across all eight metrics, with overall node-level, hierarchical, and spatial F1 scores of 89.4, 85.1, and 88.0, respectively.
These scores exceed those of GPT-6 Astra, the strongest baseline, by 13.4, 34.8, and 15.5, respectively.
Qualitative analysis further highlights two common baseline errors that our model better handles: inaccurate element localization and incorrect grouping.
For localization, Fig.~\ref{fig:qualitative}(a) shows that GPT-6 Astra shifts the bounding box of one Google ``G'' logo upward, whereas our model localizes it accurately. 
For grouping, Fig.~\ref{fig:qualitative}(b) shows that GPT-6 Astra incorrectly groups elements associated with the U.S. and Canada under the same parent, 
whereas our model correctly assigns them to separate country groups.
These results show that \dataset{} provides effective supervision for image-to-scene-graph prediction.


\subsection{Scene-Graph-Augmented Infographic QA}


\subsubsection{Benchmark Construction}
To better assess the structural reasoning capabilities of current LVLMs, we construct a QA benchmark that requires resolving associations among visual elements.
The benchmark is built in two stages: manual question construction and model-assisted expansion.
In the first stage, we select complex infographic charts from \dataset{} and manually create question-answer pairs that require
1) \textbf{resolving correspondences among elements within a group}, such as matching labels and values to their corresponding data marks;
2) \textbf{reasoning across groups}, such as identifying which country has a given value in one year and retrieving its value in another year; or
3) \textbf{reasoning about relationships between elements}, such as determining which nodes are connected or share a boundary.
This process results in 200 question-answer pairs.
In the second stage, GPT-6 Astra~\citep{openai2026gpt6astra} uses these pairs to derive an agent skill for identifying additional infographic charts and generating question-answer pairs that require similar forms of structural reasoning.
Each generated question-answer pair is verified by three authors.
After verification, the resulting benchmark contains 1,266 questions across 1,000 infographic charts.
We evaluate model accuracy using the answer-type-specific scoring rules from InfoChartQA~\citep{lin2025infochartqa}.
Example question-answer pairs and details of the scoring are provided in Appendices~\ref{app:qa-examples} and \ref{app:qa-metric}.

\subsubsection{Experimental Setup}
\mypara{Scene-graph-augmented prompting}
In scene-graph-augmented prompting, we provide scene graph annotations alongside the chart image and question to facilitate evidence identification and reasoning.
Specifically, we instruct the model to use the annotations to 1) identify relevant labels, data marks, and values;
2) establish their correspondences; and 
3) perform the necessary comparison or calculation to derive the answer.
To reduce unnecessary detail, we summarize each node's description into a concise label and remove the color palettes before providing the scene graph to the model.

\mypara{Baseline prompting methods}
We compare our method with two baselines:
1) \textbf{Direct prompting}, which asks the model to answer the question directly from the chart image; and 
2) \textbf{Evidence-guided prompting}, which follows the same identification and reasoning process as our method but without access to the scene graph annotations.
Please refer to Appendix~\ref{app:qa-prompts} for the detailed prompts.

\subsubsection{Results and Analysis}
Table~\ref{tab:vqa_models} compares QA accuracy under scene-graph-augmented prompting and the two baselines across the same ten LVLMs as in Sec.~\ref{subsec:i2jmodel}. 
Scene-graph-augmented prompting consistently outperforms both baselines across all ten models.
Compared with evidence-guided prompting, it improves accuracy by an average of 8.9\%.
The largest improvement is observed for Gemma 4-12B-IT, whose accuracy increases by 27.2\%, from 40.2\% to 67.4\%.
These results show that explicitly providing structural information can improve QA performance by reducing the need for models to infer element associations directly from the image.

\begin{table}[!t]
\centering
\caption{Comparison of QA accuracy under different prompting methods.}
\label{tab:vqa_models}
\renewcommand{\arraystretch}{1.3}
\resizebox{\linewidth}{!}{%
\begin{tabular}{l|ccccccc|ccc}
\hline
& \multicolumn{7}{c|}{Open-source models}
& \multicolumn{3}{c}{Proprietary models} \\
Prompting
& \shortstack{Molmo2\\8B}
& \shortstack{InternVL3.5\\8B}
& \shortstack{Gemma 4\\12B-IT}
& \shortstack{Qwen3.5\\4B}
& \shortstack{Qwen3.8\\27B}
& \shortstack{Qwen3.8\\Flash-Next\\FP8}
& \shortstack{GLM-5.3\\Flash}
& \shortstack{Gemini 3.8\\Flash}
& \shortstack{GPT-5.6\\Sol}
& \shortstack{GPT-6\\Astra} \\
\hline
Direct
& 37.5 & 31.9 & 32.4 & 35.9 & 75.1
& 72.7 & 74.0 & 90.0 & 81.4 & 95.9 \\
Evidence-guided
& 40.4 & 33.8 & 40.2 & 46.7 & 76.9
& 76.0 & 79.0 & 88.5 & 85.1 & 94.0 \\
\hline
\rowcolor{blue!10} 
Scene-graph-augmented
& \textbf{46.7} & \textbf{45.3} & \textbf{67.4}
& \textbf{53.4} & \textbf{86.9} & \textbf{86.1}
& \textbf{83.9} & \textbf{90.6} & \textbf{91.6} & \textbf{97.6} \\
\hline
\end{tabular}%
}
\end{table}

\begin{wraptable}[14]{r}{0.47\textwidth}
\vspace{-2.0em}
\centering
\small
\setlength{\tabcolsep}{4pt}
\caption{QA accuracy on Qwen3.8-27B with different scene graph inputs.}
\label{tab:vqa_ablation}
\renewcommand{\arraystretch}{1.15}
\begin{tabular}{l|c}
\hline
Annotations & Accuracy \\
\hline
No scene graph & 76.9 \\
\quad \makecell[l]{+ node types\ \& \\\hphantom{+~}bounding boxes} & 77.8 \\
\quad + relationships & 78.9 \\
\quad + labels & 80.2 \\
\quad + text without digits & 81.5 \\
\quad + text with digits (best) & \textbf{86.9} \\
\hline
Best + color palettes & 86.5 \\
Best, labels $\rightarrow$ descriptions & 86.5 \\
Best - relationships & 85.6 \\
\hline
\end{tabular}
\end{wraptable}
To examine the contributions of each scene graph component, we conduct an ablation study on Qwen3.8-27B.
Starting from evidence-guided prompting without scene graph annotations, we progressively add node types and bounding boxes, relationships, concise labels, text without digits, and text with digits.
As shown in Table~\ref{tab:vqa_ablation}, accuracy improves as additional scene graph components are introduced.
The largest gain comes from adding text with digits, indicating that explicit numerical text is particularly useful for answering questions that involve data values.
We further evaluate two alternative representations: adding color palettes and replacing concise labels with full descriptions.
Both slightly reduce accuracy, supporting our choice to remove color palettes and use concise labels that focus on what each element represents.
Removing relationships also reduces accuracy, suggesting that explicit relationship information can help the model establish correspondences among elements.

\subsection{Scene Graph Preservation in Image Generation}


\subsubsection{Experimental Setup}
\mypara{Models}
We evaluate how well existing open-source and proprietary image-generation models preserve specified elements and relationships when generating from scene graphs.
Specifically, we select seven representative open-source models (Qwen-Image-2512~\citep{wu2025qwenimage}, ERNIE-Image-Turbo~\citep{liu2026ernieimage}, SenseNova-U1 Infographic V3~\citep{diao2026sensenovau1}, GLM-Image~\citep{zai2026glmimage}, HiDream-O1-Image~\citep{cai2026hidream}, Ideogram 4~\citep{ideogram2026ideogram4}, and Qwen-Image-SP~\citep{chen2026scaling}) and four proprietary ones (GPT-Image-2~\citep{openai2026gptimage2}, GPT-Image-2.5 Sunburst~\citep{openai2026gptimage25}, Nano Banana Pro~\citep{google2025nanobananapro}, and Nano Banana 2~\citep{google2026nanobanana2}).
For models that support JSON inputs (Qwen-Image-SP and Ideogram 4), we convert the scene graph into the format required by each model.
For the remaining models, we provide the scene graph directly in the text prompt.
Details of the model inputs are provided in Appendix~\ref{app:generation}.\looseness=-1

\mypara{Evaluation metrics}
To measure scene graph preservation, we extract a scene graph from each generated image using our annotation pipeline and compare it with the input scene graph.
We use the same node- and edge-level metrics as in the image-to-scene-graph model evaluation.

\subsubsection{Results and analysis}
Table~\ref{tab:sg_preservation} compares scene graph preservation of the evaluated models on the \dataset{} test set.
Overall, the proprietary models perform better than the open-source ones, particularly in preserving hierarchical and spatial relationships.
GPT-Image-2.5 Sunburst achieves the highest overall node-level, hierarchical, and spatial F1 scores of 69.3, 75.4, and 64.2, respectively.
Among the open-source models, Ideogram 4 performs the best, achieving an overall node-level F1 score of 63.1 and even the highest bounding-box similarity of 47.4 across all models.
However, its hierarchical and spatial F1 scores of 58.5 and 49.2 remain substantially below those of GPT-Image-2.5 Sunburst.
This is because Ideogram 4 is trained to generate images from JSON input that specifies individual elements and their bounding boxes but does not explicitly encode relationships among them~\citep{ideogram2026ideogram4}.

\begin{figure}[!t]
\centering
\includegraphics[width=\linewidth]{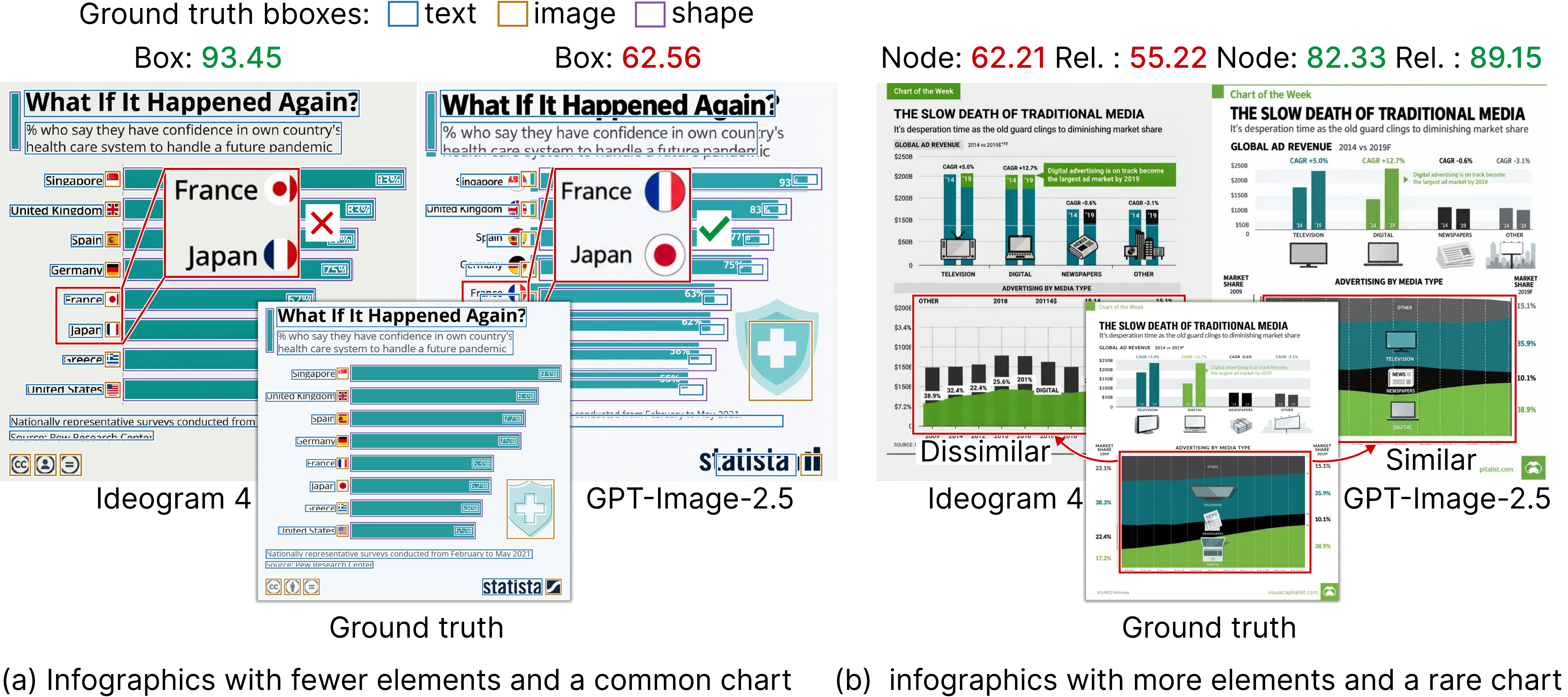}
\caption{Qualitative comparison of Ideogram 4 and GPT-Image-2.5 Sunburst.}
\label{fig:qualitative_j2i}
\end{figure}

To further examine the gap between the strongest open-source and proprietary models, we qualitatively compare Ideogram 4 and GPT-Image-2.5 Sunburst.
According to our observation, for infographics with fewer elements and common chart types, Ideogram 4 better preserves the specified layout but is more prone to errors in element content, whereas GPT-Image-2.5 Sunburst better preserves element content but exhibits larger layout deviations.
For example, in Fig.~\ref{fig:qualitative_j2i}(a), Ideogram 4 closely follows the specified layout but misrenders the French and Japanese flags, while GPT-Image-2.5 Sunburst renders them correctly but at inaccurate positions.
For infographics with more elements and rare chart types, GPT-Image-2.5 Sunburst shows a clearer advantage in preserving both elements and their relationships (Fig.~\ref{fig:qualitative_j2i}(b)).

\begin{table}[!t]
\centering
\caption{Comparison of scene graph preservation across representative image-generation models.}
\label{tab:sg_preservation}
\renewcommand{\arraystretch}{1.2}
\resizebox{\linewidth}{!}{%
    \begin{tabular}{l|c|ccccc|cc}
        \hline
        & \multicolumn{6}{c|}{Node-level} 
        & \multicolumn{2}{c}{Edge-level} \\
        \hline
        Model 
        & Overall
        & Type
        & Box
        & Text
        & Color
        & Description
        & Hierarchical 
        & Spatial \\
        \hline
        HiDream-O1-Image              & 18.7 & 92.1 & 32.8 & 20.8 & 75.9 & 40.1 & 16.6 & 6.6 \\
        GLM-Image                     & 35.9 & 83.7 & 25.0 & 19.1 & 74.3 & 35.9 & 20.8 & 14.0 \\
        Qwen-Image-2512               & 40.1 & 82.5 & 28.2 & 15.9 & 77.8 & 36.4 & 28.0 & 19.1 \\
        Qwen-Image-SP                 & 44.9 & 85.9 & 31.7 & 24.9 & 78.7 & 38.1 & 35.2 & 25.3 \\
        ERNIE-Image-Turbo             & 45.3 & 89.6 & 30.3 & 34.2 & 78.6 & 45.1 & 36.9 & 23.2 \\
        SenseNova-U1 Infographic V3   & 48.5 & 92.5 & 30.9 & 60.7 & 77.9 & 54.6 & 40.7 & 26.0 \\
        Ideogram 4                    & 63.1 & 93.0 & \textbf{47.4} & 67.4 & 80.5 & 57.8 & 58.5 & 49.2 \\
        \hline
        Nano Banana 2                 & 61.7 & 93.5 & 38.1 & 85.6 & 80.9 & 66.4 & 64.8 & 48.2 \\
        Nano Banana Pro               & 65.4 & 94.1 & 39.2 & \textbf{89.4} & 85.0 & 68.4 & 70.4 & 53.5 \\
        GPT-Image-2                   & \underline{68.5} & \underline{95.3} & 39.0 & \underline{86.2} & \underline{88.2} & \underline{69.4} & \underline{74.4} & \underline{62.5} \\
        GPT-Image-2.5 Sunburst        & \textbf{69.3} & \textbf{95.6} & \underline{43.1} & \underline{86.2} & \textbf{89.1} & \textbf{70.3} & \textbf{75.4} & \textbf{64.2} \\
        \hline
    \end{tabular}%
}
\end{table}

\section{Conclusion}
\label{sec:conclusion}

In this paper, we present \dataset{}, a richly annotated dataset for structure-aware chart understanding and generation.
It includes \totalnum{} real and synthetic infographic charts, each paired with a scene graph that captures the chart's visual elements, as well as their hierarchical and spatial relationships.
Through three applications, we demonstrate the utility of \dataset{} for developing and evaluating models that extract, use, and preserve the structure of infographic charts.
Despite these promising results, several directions remain for future work.
First, representing element locations with segmentation masks or vector paths, rather than bounding boxes, could better capture element shapes.
Second, directly training image-generation models with scene graph annotations could further explore their value for structure-aware generation.

\section*{AI Use Statement}


In this work, we used generative AI tools for generating synthetic datasets, implementing methods, and cleaning and reformatting datasets.
We have not used generative AI tools for proposing or refining hypotheses, designing or providing feedback on research methodology or experiments, assisting with translation, supporting qualitative and thematic data analysis, or interpreting results.
Developing theoretical models or conceptual frameworks, formulating mathematical claims, providing critical ingredients for proving mathematical claims, and assisting in the writing of proofs are not applicable to this work.
Additionally, we used generative AI tools for brainstorming and sourcing or searching for information. 
We have reviewed all AI-assisted work. 
LLM-generated code for extracting scene graph annotations was manually reviewed by the authors for validity.
We also verified the sources of all information provided by LLMs.
We take responsibility for the final content of this work, including text, claims, or artifacts produced with the aid of generative AI.


\section*{Ethics Statement}
We carefully consider copyright and licensing issues in the construction and release of \dataset{}.
To avoid redistributing third-party content, we release only the source URLs of the real infographic chart images together with our newly constructed scene graph annotations, and do not host or redistribute the original images.
We have reviewed the licenses and reuse conditions of the source platforms and confirmed that our use and release protocol is consistent with these terms.
Further details on licenses, reuse conditions, and attribution requirements are provided in Appendix~\ref{app:dataset-release}.

\bibliography{iclr2027_conference}
\bibliographystyle{iclr2027_conference}

\newpage
\appendix

\section{Scene Graph Schema}
\label{app:schema}

This section specifies the node attributes, hierarchical relationships, and spatial relationships introduced in Sec.~\ref{sec:scene_graph}, together with their JSON representation.
The hierarchy records which elements jointly form semantic components, while spatial relationships describe the geometry of their bounding boxes.

\subsection{Node Attribute Specification}
\label{app:node-attributes}

The field \texttt{node\_kind} distinguishes individual visual elements (\texttt{element}) from semantic groups (\texttt{group}).
An element has one of three types: \texttt{text} for a visible text region, \texttt{image} for an image object such as an icon, photograph, or illustration, and \texttt{shape} for a geometric primitive such as a mark, line, or border.
Group types identify the semantic components described below.
The corresponding type is stored in \texttt{element\_type} or \texttt{group\_type}, respectively.

Every stored node has a bounding box, color palette, and description, in addition to its type and structural identifiers (Table~\ref{tab:app-node-fields}).
Bounding boxes use integer coordinates in the order $[y_0,x_0,y_1,x_1]$, corresponding to the top, left, bottom, and right boundaries.
The image origin is at the top-left corner, and horizontal and vertical coordinates are normalized independently to $[0,1000]$ using the image width and height.
Boxes have positive width and height: $x_0<x_1$ and $y_0<y_1$.
This convention applies to both elements and groups.

\begin{table}[!htbp]
\centering
\caption{Node fields in the JSON representation. Fields marked ``element'' or ``group'' apply only to that node kind.}
\label{tab:app-node-fields}
\small
\renewcommand{\arraystretch}{1.15}
\begin{tabular}{@{}p{0.24\linewidth}p{0.13\linewidth}p{0.56\linewidth}@{}}
\toprule
Field & Applies to & Definition \\
\midrule
\texttt{label} & All & Unique node identifier within the image, combining its type and one-based position in the node array, e.g., \texttt{<text 4>}. \\
\texttt{parent} & All & Label of the direct parent group, or the reserved token \texttt{ROOT}. \\
\texttt{node\_kind} & All & \texttt{element} or \texttt{group}. \\
\texttt{element\_type} & Element & \texttt{text}, \texttt{image}, or \texttt{shape}. \\
\texttt{group\_type} & Group & One of the seven semantic group types. \\
\texttt{bbox} & All & Four integers $[y_0,x_0,y_1,x_1]$ in the range $[0,1000]$. \\
\texttt{color\_palette} & All & Array of distinct dominant colors, represented as \texttt{\#RRGGBB} or \texttt{\#RRGGBBAA}; the final two digits encode alpha. \\
\texttt{desc} & All & Nonempty natural-language description of the node's visible appearance, composition, and functional role. \\
\texttt{text} & Text element & Exact visible wording, preserving case, spelling, numbers, punctuation, and line breaks. \\
\bottomrule
\end{tabular}
\end{table}

The \texttt{text} field preserves the visible string.
The \texttt{desc} field describes the node's visible appearance, composition, and functional role within the infographic.
For example, the description of a \texttt{shape} can indicate that it serves as a data mark, a legend key, or a decorative border.
For text regions, it describes typography, arrangement, and function, while the exact wording is stored in \texttt{text}.
Group descriptions explain the component formed by their members; group nodes do not carry a separate text-transcription field.

\subsection{Hierarchical Relationships}
\label{app:grouping}

Hierarchical edges point from a parent group to its children.
In JSON, each child stores its incoming hierarchical edge through its \texttt{parent} field; a separate list of hierarchical edges is therefore unnecessary.
Every stored node has exactly one parent, which must be an existing group or \texttt{ROOT}, and every parent chain terminates at \texttt{ROOT} without a cycle.
The hierarchy thus forms a rooted tree when the implicit root is included.
\texttt{ROOT} represents the whole image and is not stored as a node object or counted as an additional ordinary group type.

Groups identify components whose members jointly convey information (Table~\ref{tab:app-group-types}).
For example, a \texttt{data\_item} can associate a mark with its category and value labels, while a \texttt{legend\_item} associates a legend key with its label.
A group can contain individual elements and other groups, allowing a chart to contain data items, axes, and a legend, with the legend further containing its entries.
Parent membership expresses semantic composition; bounding-box containment or spatial proximity alone does not establish this relationship.

\begin{table}[!htbp]
\centering
\caption{Group types and their intended meanings.}
\label{tab:app-group-types}
\small
\renewcommand{\arraystretch}{1.15}
\begin{tabular}{@{}p{0.19\linewidth}p{0.76\linewidth}@{}}
\toprule
Group type & Meaning \\
\midrule
\texttt{chart} & A chart component containing data marks or data items and their associated reading references. \\
\texttt{axis} & An axis and its associated components, such as ticks, tick labels, and an axis title. \\
\texttt{legend} & A legend containing its entries and any associated title. \\
\texttt{legend\_item} & One legend entry associating a visual key with its label. \\
\texttt{data\_item} & Elements jointly representing one observation, such as a mark with its category label, value, or identifying icon. \\
\texttt{series} & Elements representing one data series, including its marks and associated labels. \\
\texttt{panel} & A coherent layout region, such as a heading block or a distinct component of a multi-part infographic. \\
\bottomrule
\end{tabular}
\end{table}

\subsection{Spatial Relationships}
\label{app:spatial-relations}

Spatial relationships are computed from node bounding boxes.
Let $B_a=(t_a,l_a,b_a,r_a)$ and $B_b=(t_b,l_b,b_b,r_b)$ denote two boxes in normalized coordinates, and let $W$ and $H$ be the original image width and height in pixels.
Each spatial record specifies its type through \texttt{relation\_type} and its endpoints through \texttt{from\_label} and \texttt{to\_label}.
Following Sec.~\ref{sec:scene_graph}, we describe four types of spatial relationships: direction, alignment, overlap, and proximity.
The selection rule below determines which node pairs are included in the serialized representation.

\paragraph{Direction.}
Direction describes where the first box lies relative to the second, using eight values: \texttt{top}, \texttt{bottom}, \texttt{left}, \texttt{right}, \texttt{top-left}, \texttt{top-right}, \texttt{bottom-left}, and \texttt{bottom-right}.
We determine the horizontal and vertical components using the conditions
\begin{equation}
\begin{aligned}
\mathrm{left}:&\quad r_a\leq l_b, &
\mathrm{right}:&\quad l_a\geq r_b,\\
\mathrm{top}:&\quad b_a\leq t_b, &
\mathrm{bottom}:&\quad t_a\geq b_b.
\end{aligned}
\end{equation}
When only a horizontal or vertical condition holds, it gives one of the four cardinal directions.
When both hold, their combination gives a diagonal direction, written with the vertical component first and joined by a hyphen, e.g., \texttt{top-left}.
Directional relationships are serialized with \texttt{relation\_type} set to \texttt{relative\_position}.
The \texttt{directions} array stores a single value, e.g., \texttt{["top-left"]}.
If neither axis is separated, no directional record is stored for the pair.

\paragraph{Alignment.}
\texttt{aligned\_with} compares corresponding left, right, top, and bottom boundaries and horizontal and vertical centers.
Writing $c_x=(l+r)/2$ and $c_y=(t+b)/2$, an alignment basis $q\in\{l,r,t,b,c_x,c_y\}$ is satisfied when $|q_a-q_b|\leq 5$ in normalized coordinates.
The \texttt{basis} array contains all satisfied bases, encoded as \texttt{left}, \texttt{right}, \texttt{top}, \texttt{bottom}, \texttt{center\_x}, or \texttt{center\_y}.
The tolerance corresponds to $0.5\%$ of image width for horizontal coordinates and $0.5\%$ of image height for vertical coordinates.

\paragraph{Overlap.}
For a selected pair, \texttt{bbox\_overlaps} is present exactly when the two boxes have a positive-area intersection:
\begin{equation}
\min(r_a,r_b)>\max(l_a,l_b)
\quad\text{and}\quad
\min(b_a,b_b)>\max(t_a,t_b).
\end{equation}
Containment is included as an overlap case, while touching boundaries alone do not satisfy this condition.

\paragraph{Proximity.}
Proximity records geometric adjacency through the Boolean field \texttt{is\_near} and also stores the distance between the bounding boxes.
The Boolean predicate is independent of the sibling-based pair selection used for sparse storage below.
The box distance is the shortest gap between the two rectangular regions, with a smaller value indicating closer boxes.
We first convert horizontal and vertical gaps to pixels:
\begin{equation}
\begin{aligned}
\Delta_x &= \frac{W}{1000}\max(0,l_a-r_b,l_b-r_a),\\
\Delta_y &= \frac{H}{1000}\max(0,t_a-b_b,t_b-b_a),\\
d_{\mathrm{px}} &= \sqrt{\Delta_x^2+\Delta_y^2},\qquad
d_{\mathrm{ratio}} = \frac{d_{\mathrm{px}}}{\sqrt{W^2+H^2}}.
\end{aligned}
\end{equation}
Proximity is serialized with \texttt{relation\_type} set to \texttt{bbox\_distance}.
Alongside \texttt{is\_near}, its fields \texttt{distance\_px} and \texttt{distance\_ratio} store these values rounded to six and ten decimal places, respectively.
Overlapping or touching boxes have zero distance.
Using $W$ and $H$ separately preserves the original image aspect ratio when computing the pixel distance.
For valid positive-area boxes, let $A_a=WH(r_a-l_a)(b_a-t_a)/10^6$ and define $A_b$ analogously, both in square pixels.
The Boolean predicate is
\begin{equation}
\begin{split}
\texttt{is\_near}(a,b) ={}& \neg\texttt{bbox\_overlaps}(a,b)\\
&{}\land [d_{\mathrm{px}}\leq 0.06\min(W,H)]\\
&{}\land [d_{\mathrm{px}}\leq \min(\sqrt{A_a},\sqrt{A_b})].
\end{split}
\end{equation}
Thus positive-area overlap, including containment, is not near, whereas exact edge or corner contact is near.
Separated pairs must satisfy both the image-scale and node-scale gap limits.
Missing or invalid boxes do not create near relationships.

\paragraph{Selection of spatial relationships.}
Exhaustive pairwise annotation would repeat layout information across many unrelated or redundant node pairs.
We therefore select relationships between direct siblings at each level of the semantic hierarchy, including children of \texttt{ROOT}.
Element-level relationships describe the arrangement within a component, while relationships between sibling groups describe the arrangement of components.
Parent membership remains represented by the hierarchical edges.

Within each sibling set, we retain the union of three kinds of node pairs: nearest-sibling candidates, consecutive nodes within aligned sets, and overlapping nodes.
Nearest-sibling candidates retain local arrangements; alignment chains describe rows, columns, and shared boundaries without requiring every aligned pair; overlap preserves intersections such as an illustration placed inside a bar.
For each retained pair, we record all applicable direction, alignment, and overlap information, together with its proximity flag and box distance.
Each relationship type is stored once per unordered pair; direction uses the specified source and target order without a duplicate inverse record.
Unrecorded pairs are not treated as negative labels, and their geometry remains available from the node bounding boxes.

For nearest-sibling candidates, each node nominates its two nearest siblings by box distance, and each group additionally nominates its two nearest sibling groups.
When fewer than two eligible nodes exist, all are nominated; ties at the distance cutoff are retained.
A pair is a nearest-sibling candidate if either endpoint nominates the other; nomination does not determine its Boolean proximity label.
For each alignment basis, we sort siblings by that coordinate and greedily form consecutive sets whose coordinate span is at most 5 normalized units.
Within each set, consecutive nodes ordered by their centers along the perpendicular axis define the alignment chain; ties are resolved by the aligned coordinate and then node label.
The bounded coordinate span prevents approximate alignment from accumulating across a chain.
All positive-area sibling overlaps are retained, and the union is deduplicated without an image-wide record cap.
These storage rules are separate from the metric computation described in Appendix~\ref{app:i2sg}.

\subsection{JSON Format Example}
\label{app:serialization}

The JSON representation of a scene graph contains four top-level fields.
The \texttt{high\_level\_description} field describes the overall image composition, and \texttt{style\_description} records its visual style.
The \texttt{compositional\_deconstruction} object contains a background description and the node array, whose \texttt{parent} references specify the hierarchy.
The \texttt{relations} array contains spatial records with the type-specific fields defined above.
Image dimensions are read from the image itself and are not included in this JSON representation.
The image-to-scene-graph training target omits \texttt{relations}, since these relationships can be computed from the predicted bounding boxes.

\begin{figure}[!ht]
\centering
\includegraphics[width=\linewidth]{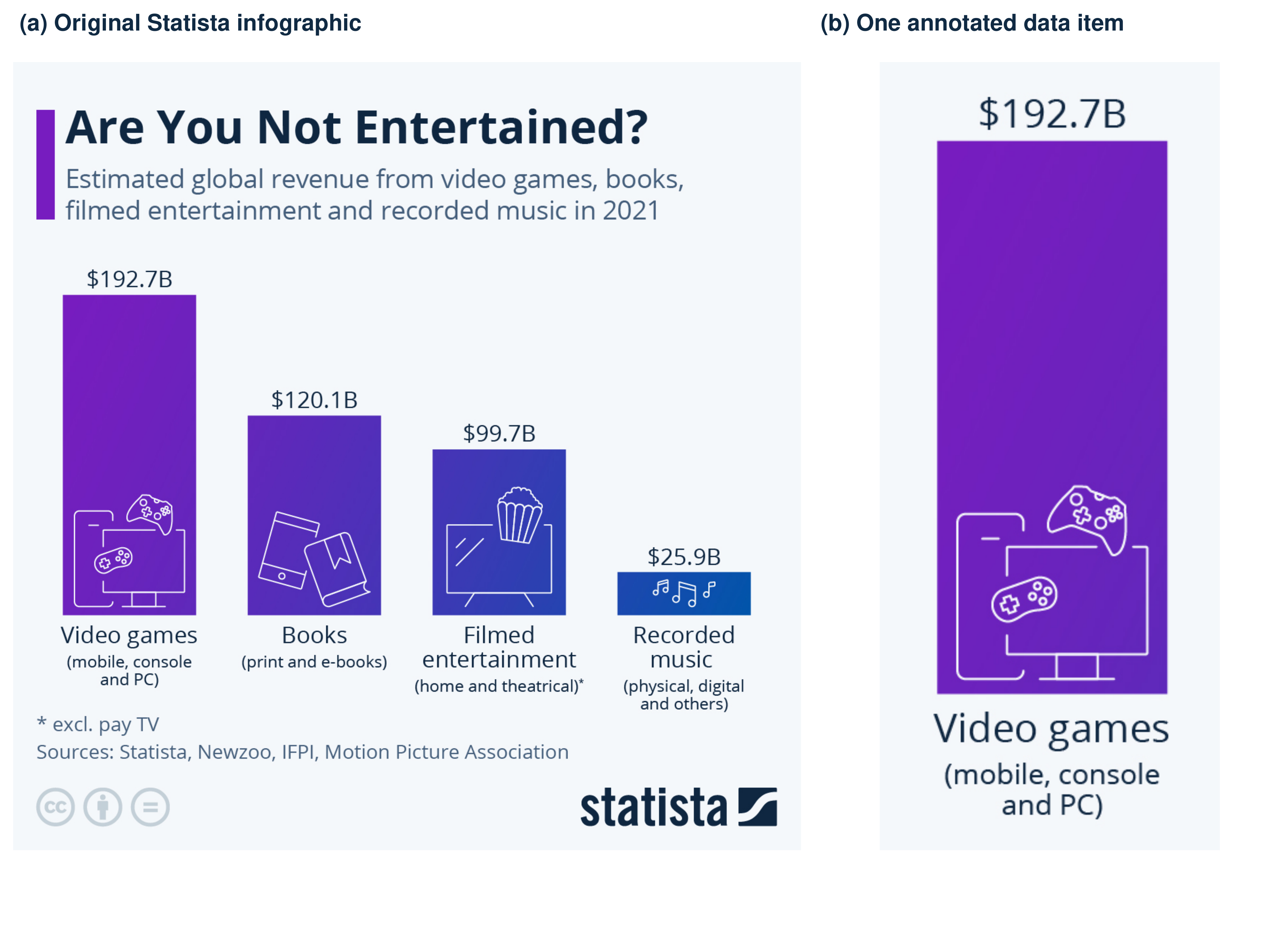}
\caption{A real infographic and its scene-graph annotation.
(a) Original four-bar Statista infographic by Felix Richter (2022), with its source and attribution retained.
(b) Enlarged view of the video-games data item: group 2 contains bar 3, illustration 4, revenue label 5, category label 6, and explanatory text 7.
The annotation boxes and numbered tags are added for this example and correspond to the JSON excerpt below.}
\label{fig:app-json-example}
\end{figure}

\newpage
Figure~\ref{fig:app-json-example} uses a published Statista infographic comparing four entertainment sectors~\citep{richter2022entertained}.
The original $1200\times1200$ image is reproduced alongside an enlarged view of its first data item with manually annotated boxes.
This data-item group associates the bar, gaming illustration, revenue label, category label, and category-coverage text.
Solid boxes denote elements, and the dashed box denotes their parent group.
Numbers match the numeric suffixes of the JSON node labels.
All box coordinates are normalized relative to the full source image.

The excerpt retains the chart node, one complete data-item subtree, and the spatial records for two pairs within that data item.
The other data items, heading, footer, and spatial records are omitted for readability.
The illustration and bar overlap, giving a box distance of zero.
The bar is \texttt{bottom} relative to its revenue label; their horizontal centers are aligned, and the normalized boxes give a gap of 8.4 pixels.
The overlapping illustration--bar pair has \texttt{is\_near: false}; the separated bar--label pair satisfies both gap limits and has \texttt{is\_near: true}.

{\fontsize{8}{9.2}\selectfont
\begin{verbatim}
{
  "high_level_description":
    "Four entertainment sectors compared by revenue.",
  "style_description": {
    "aesthetics": "Clean editorial infographic.",
    "lighting": "Uniform.", "photo": null, "medium": "graphic_design",
    "art_style": "Flat vector with gradient bars.",
    "color_palette": ["#F3F8FC", "#102A43", "#7420BB",
                      "#0756AC", "#FFFFFF"]
  },
  "compositional_deconstruction": {
    "background": "Pale blue background.",
    "nodes": [
      {"label": "<chart 1>", "parent": "ROOT",
       "node_kind": "group", "group_type": "chart",
       "bbox": [261, 56, 826, 938],
       "color_palette": ["#102A43", "#7420BB", "#0756AC"],
       "desc": "Four bars with category icons and labels."},
      {"label": "<data_item 2>", "parent": "<chart 1>",
       "node_kind": "group", "group_type": "data_item",
       "bbox": [261, 56, 794, 237],
       "color_palette": ["#7420BB", "#FFFFFF", "#102A43"],
       "desc": "Bar, icon, value, and category labels."},
      {"label": "<shape 3>", "parent": "<data_item 2>",
       "node_kind": "element", "element_type": "shape",
       "bbox": [295, 63, 703, 234],
       "color_palette": ["#7420BB"],
       "desc": "Vertical gradient bar encoding revenue."},
      {"label": "<image 4>", "parent": "<data_item 2>",
       "node_kind": "element", "element_type": "image",
       "bbox": [547, 76, 695, 221],
       "color_palette": ["#FFFFFF"],
       "desc": "White gaming devices used as a category icon."},
      {"label": "<text 5>", "parent": "<data_item 2>",
       "node_kind": "element", "element_type": "text",
       "bbox": [261, 91, 288, 205],
       "color_palette": ["#102A43"],
       "desc": "Navy revenue label above the bar.",
       "text": "$192.7B"},
      {"label": "<text 6>", "parent": "<data_item 2>",
       "node_kind": "element", "element_type": "text",
       "bbox": [713, 56, 744, 237],
       "color_palette": ["#102A43"],
       "desc": "Navy category label below the bar.",
       "text": "Video games"},
      {"label": "<text 7>", "parent": "<data_item 2>",
       "node_kind": "element", "element_type": "text",
       "bbox": [750, 66, 794, 230],
       "color_palette": ["#102A43"],
       "desc": "Navy text specifying category coverage.",
       "text": "(mobile, console\nand PC)"}
    ]
  },
  "relations": [
    {"relation_type": "aligned_with",
     "from_label": "<image 4>", "to_label": "<shape 3>",
     "basis": ["center_x"]},
    {"relation_type": "bbox_overlaps",
     "from_label": "<image 4>", "to_label": "<shape 3>"},
    {"relation_type": "bbox_distance",
     "from_label": "<image 4>", "to_label": "<shape 3>",
     "is_near": false, "distance_px": 0.0,
     "distance_ratio": 0.0},
    {"relation_type": "relative_position",
     "from_label": "<shape 3>", "to_label": "<text 5>",
     "directions": ["bottom"]},
    {"relation_type": "aligned_with",
     "from_label": "<shape 3>", "to_label": "<text 5>",
     "basis": ["center_x"]},
    {"relation_type": "bbox_distance",
     "from_label": "<shape 3>", "to_label": "<text 5>",
     "is_near": true, "distance_px": 8.4,
     "distance_ratio": 0.0049497475}
  ]
}
\end{verbatim}
}

\section{Annotation Pipelines and Quality Assessment}
\label{app:annotation}

\subsection{Real Infographic Annotation}
\label{app:annotation-prompts}

Following Sec.~\ref{sec:annotation}, the image-based pipeline constructs a scene graph through element annotation and relationship extraction, followed by iterative refinement.
Element annotation identifies the visible objects and their attributes; relationship extraction organizes these elements into a semantic hierarchy and computes their spatial relationships.
Table~\ref{tab:app-annotation-stages} summarizes the information passed between stages.
We describe validation in Appendix~\ref{app:annotation-refinement} and provide the prompt templates in Appendix~\ref{app:annotation-templates}.
The same image-based procedure applies when annotating generated infographics; it uses the image being annotated, without access to a paired reference image or its scene graph.

\paragraph{Element annotation.}
HunyuanOCR-1.5 provides text content and bounding boxes, while DAD provides candidate bounding boxes for non-text elements.
We normalize these boxes to the $[y_0,x_0,y_1,x_1]$ convention in Appendix~\ref{app:node-attributes}.
Given the original image, contextual crops, and localized candidates, Gemini 3.8 Flash annotates each element's type, color palette, and description, while retaining the supplied text and geometry at this stage.
The description includes the element's functional role, such as a data mark, category label, legend key, or decoration.
The resulting elements form the leaf nodes used by relationship extraction; visual errors in localization or attributes are handled by the refinement stage.

\begin{table}[!htbp]
\centering
\caption{Inputs and outputs of the image-based annotation pipeline. Hierarchical relationships are inferred from the image and annotated elements; spatial relationships are computed from node geometry.}
\label{tab:app-annotation-stages}
\small
\renewcommand{\arraystretch}{1.15}
\begin{tabular}{@{}>{\raggedright\arraybackslash}p{0.25\linewidth}>{\raggedright\arraybackslash}p{0.30\linewidth}>{\raggedright\arraybackslash}p{0.38\linewidth}@{}}
\toprule
Stage & Input & Output \\
\midrule
Element annotation & Image, crops, and localized candidates & Leaf nodes with type, box, palette, description, and text where applicable. \\
Hierarchical relationship extraction & Image and annotated elements & Semantic groups and parent--child assignments. \\
Spatial relationship extraction & Element and group boxes & Direction, alignment, overlap, and proximity relationships. \\
Iterative refinement & Image, current graph, and validation findings & Corrected annotations, rechecked graph, and remaining unresolved findings. \\
\bottomrule
\end{tabular}
\end{table}

\paragraph{Hierarchical relationship extraction.}
Gemini 3.8 Flash receives the original image and the annotated elements and identifies components whose members jointly convey information.
The model uses visible region boundaries, relative placement, and semantic associations to identify chart groups and associate marks with their labels and values.
For example, a country name, flag, bar, and value label can jointly represent one data item.
Shared components, such as axes and legends, are represented by the corresponding group types in Appendix~\ref{app:schema}.

The grouping prompt specifies the intended annotation granularity, including what constitutes a data item or continuous series and how shared chart context is represented.
These conventions make group meanings consistent across annotations, while the model determines the memberships from each image.
The element inventory is fixed during this stage, and the model returns a structured plan specifying semantic groups and element assignments.
The program converts that plan into internal nodes and hierarchical edges, with one direct parent per element.
Group boxes enclose their descendant elements, group palettes combine the descendants' colors, and group descriptions summarize the resulting components.

\paragraph{Spatial relationship extraction.}
We compute direction, alignment, overlap, and proximity from node bounding boxes using the definitions and relationship-selection rules in Appendix~\ref{app:spatial-relations}.
This computation applies to the elements and groups in the constructed hierarchy and requires no additional model request.
After refinement changes node geometry or group membership, the affected group boxes and spatial relationships are recomputed.

\paragraph{Model and request configuration.}
Table~\ref{tab:app-annotation-settings} gives the model settings for element annotation, relationship extraction, and refinement.
Gemini requests use JSON response mode; temperature and top-$p$ are left unspecified in the request.
Output schemas are checked by the program, including when they are also passed through the API's structured-output parameter.
Spatial relationships are computed programmatically, so they require no additional prompt.

\begin{table}[!htbp]
\centering
\caption{Model settings for the image-based annotation pipeline. Token limits are per request.}
\label{tab:app-annotation-settings}
\small
\renewcommand{\arraystretch}{1.12}
\begin{tabular}{@{}>{\raggedright\arraybackslash}p{0.28\linewidth}>{\raggedright\arraybackslash}p{0.67\linewidth}@{}}
\toprule
Component & Setting \\
\midrule
HunyuanOCR-1.5 & \texttt{spotting\_json}; temperature 0, top-$p$ 1, repetition penalty 1.08; at most 32,768 output tokens. \\
DAD & Fixed layer-localization instruction; temperature 0, top-$p$ 0.9, top-$k$ 20; deployed output limit 16,384 tokens. \\
Image and crop inputs & Up to $3\times3$ ownership cores; grid size depends on image dimensions and candidate count, with a 900-pixel size threshold and 80-pixel context padding. \\
Element attributes / scene description & Gemini 3.8 Flash; at most 64 candidates per attribute request; LOW thinking; at most 32,768 output tokens. \\
Localization recovery / crop correction & Gemini 3.8 Flash; LOW thinking; at most 32,768 output tokens. \\
Whole-image review & Gemini 3.8 Flash; MEDIUM thinking; at most 32,768 output tokens; at most two audit rounds in the recorded extraction configuration. \\
Semantic grouping & Gemini 3.8 Flash; LOW thinking; 16,384 output tokens, increased to 32,768 when there are more than 350 elements. \\
Grouping review & Gemini 3.8 Flash; MEDIUM thinking; at most 16,384 output tokens; initial construction/review budget of three rounds. \\
\bottomrule
\end{tabular}
\end{table}

\subsection{Validation and Refinement}
\label{app:annotation-refinement}
We validate the graph structurally and compare its annotations with the source image.
Detected errors are routed to element annotation or relationship extraction, and the updated graph is checked again.
The two forms of validation address different requirements: a valid graph structure and a faithful representation of visible content.

\paragraph{Programmatic checks.}
The program verifies required attributes, coordinate ranges, unique node identifiers, valid parent references, and an acyclic hierarchy terminating at \texttt{ROOT}.
It also checks that the grouping plan accounts for every supplied element exactly once.
Responses that fail these checks receive the validation findings as feedback for correction.

\paragraph{LVLM-based verification.}
Gemini 3.8 Flash compares the graph with the original image to identify missing or extra elements, incorrect attributes or boxes, and incorrect groupings.
Whole-image inspection checks the complete scene, while contextual crops support closer inspection of individual regions.
For hierarchical relationships, a separate review request examines the image, fixed elements, and proposed grouping plan under the same annotation conventions.
It assesses chart boundaries, the completeness of data items or series, and element membership; the program verifies that the review covers every proposed unit.

\paragraph{Iterative correction.}
Element-level findings are returned to the annotation stage, which produces additions, removals, or updates while preserving unaffected nodes.
Grouping findings are returned to relationship extraction to revise the affected assignments.
After an element correction, the affected grouping is rebuilt and reviewed, and group geometry and spatial relationships are recomputed.
The updated graph must pass structural checks and image-based verification again.
The process stops when verification identifies no further actionable errors.
The prompt templates for these requests are provided in Appendix~\ref{app:annotation-templates}.

\subsection{Annotation Quality Assessment}
\label{app:annotation-quality}
The scene graph annotations are assessed by three co-authors, denoted Expert A, Expert B, and Expert C.

\paragraph{Evaluation Dataset.}
The training and test splits are annotated using the same pipeline.
Human verification serves both to correct the test annotations for reliable evaluation and to assess the quality of the shared annotation pipeline by measuring errors before human correction.
We manually verify the scene graph annotations in our held-out test set, comprising 500 real and 500 synthetic images across three complexity levels (Table~\ref{tab:app-audit-sample}).
Complexity counts element and group nodes, including the root, before audit corrections.
We divide charts into low ($\leq50$ nodes), medium (51--91 nodes), and high ($\geq92$ nodes) complexity, with cutoffs based on the full dataset's node-count distribution.
For quality assessment, the test set contains 99,010 non-root nodes: 82,042 elements and 16,968 groups.

\begin{table}[!htbp]
\centering
\caption{Quality-assessment sample by scene graph complexity.}
\label{tab:app-audit-sample}
\small
\begin{tabular}{@{}lr@{}}
\toprule
Complexity & Images \\
\midrule
Low ($\leq50$ nodes) & 191 \\
Medium (51--91 nodes) & 344 \\
High ($\geq92$ nodes) & 465 \\
\midrule
Total & 1000 \\
\bottomrule
\end{tabular}
\end{table}

\paragraph{Review system and procedure.}
We developed an annotation review system that links the source image, bounding-box overlays, scene graph hierarchy, and node attributes.
It supports zoomed image inspection, node-level assessments and corrections, separate reviewer records, and comparison of disagreements for adjudication.
The experts use programmatic image-inspection tools and structured review records to carry out this workflow.

Experts A and B independently inspect the whole image for missing elements, then check each annotated node for text or numeric errors, incorrect boxes, semantic description errors, and incorrect parent--child relationships.
Descriptions include the element's functional role.
After both reviews are complete, Expert C compares their findings with the image, resolves disagreements, and checks for mistakes shared by both reviewers.
A finding is accepted only when supported by visible evidence; otherwise, it is rejected as unsupported or an acceptable alternative, or retained as unresolved.
Equivalent grouping choices, small box-padding differences, color approximations, and paraphrases do not count as errors.

\paragraph{Assessment measures.}
We summarize findings at the element or node level, using the relevant annotations as denominators.
For element completeness, precision is $M/N_{\mathrm{orig}}$ and recall is $M/N_{\mathrm{ref}}$: $N_{\mathrm{orig}}$ counts original elements, $M$ counts retained original elements, and $N_{\mathrm{ref}}$ equals $M$ plus confirmed missing elements.
Element correspondence follows the objects identified during review; accepted additions and removals account for omissions and extra elements.
For localization, we compute mean intersection-over-union (IoU) between corresponding original and adjudicated element boxes.
The reference incorporates supported corrections while preserving acceptable original representations; boxes without a supported reference are excluded from IoU.

We count transcription errors among retained text nodes, and numeric transcription errors among those containing digits.
Semantic description errors and incorrect assignments are counted over non-root nodes and their parent links, respectively.
Each affected node counts once per measure; a node can have several problems, while dependent updates to ancestor summaries are not counted as additional defects.
Unresolved judgments remain separate from confirmed errors.

\begin{table}[!htbp]
\centering
\caption{Node-level assessment before audit corrections. Error rows report affected nodes / eligible nodes. Numeric transcription errors are a subset of text errors. Precision, recall, and IoU are computed against the expert-adjudicated reference.}
\label{tab:app-audit-findings}
\small
\setlength{\tabcolsep}{4pt}
\begin{tabular}{@{}lr@{}}
\toprule
Measure & Result \\
\midrule
Non-root nodes & 99,010 \\
Element nodes & 82,042 \\
Element precision & 99.81\% \\
Element recall & 99.78\% \\
Mean box IoU & 0.9976 \\
Text transcription errors & 134/34,242 \\
Numeric transcription errors & 27/21,856 \\
Description / role errors & 3,053/99,010 \\
Incorrect parent assignments & 10/99,010 \\
\bottomrule
\end{tabular}
\end{table}

\paragraph{Assessment results.}
Table~\ref{tab:app-audit-findings} summarizes the node-level assessment.
The audit gives an element precision of 99.81\% and a recall of 99.78\%.
Localization corrections affect 645 of 82,042 element boxes; mean IoU is 0.9976.
Transcription errors affect 134/34,242 retained text nodes (0.39\%). Semantic descriptions or functional roles require correction in 3,053/99,010 nodes (3.08\%).

Description and role errors often concern an object's meaning even when its box and text are correct.
For example, one chart describes 98 decorative dots as data observations, while 222 nodes across four charts describe monetary values, percentages, or ratios as ranks.
Some group-summary errors arise from fixed templates and summaries left unchanged after child-node corrections.
An axis-description template mentions an axis title without checking whether one is present; in a year-mismatch case, a child's description was corrected from 1997 to 1999, but its parent summary retained 1997.
The count therefore reflects repeated semantic mistakes across individual nodes, with a single misinterpretation affecting many nodes in one chart.
These findings motivate checking descriptions against the actual visual encoding and checking group summaries against their children; wording preferences alone are excluded.

Overall, the audit results support the high quality and reliability of annotations in the test set, particularly in element completeness, localization, and text transcription.

\subsection{Prompt Templates}
\label{app:annotation-templates}
\definecolor{annotationpromptaccent}{HTML}{31556F}
\definecolor{annotationpromptbackground}{HTML}{F4F7FB}
\newtcolorbox[auto counter,number within=section]{annotationpromptbox}[2][]{
  enhanced,breakable,
  colback=annotationpromptbackground,colframe=black!65,
  colbacktitle=annotationpromptaccent!12!white,coltitle=black,
  fonttitle=\small\bfseries,
  boxrule=0.5pt,arc=0pt,outer arc=0pt,
  left=9pt,right=9pt,top=7pt,bottom=7pt,
  before skip=10pt,after skip=10pt,
  title={Prompt~\thetcbcounter. #2},
  title after break={Prompt~\thetcbcounter. #2 (continued)},#1}
\lstdefinestyle{annotationprompttext}{
  basicstyle=\ttfamily\fontsize{8.5}{10.25}\selectfont,
  columns=fullflexible,keepspaces=true,showstringspaces=false,
  breaklines=true,breakatwhitespace=true,breakindent=0pt,
  aboveskip=0pt,belowskip=0pt,
  emph={IMAGE_SIZE,TILE,CANDIDATES,SCENE_SCHEMA,GROUPING_POLICY,
    ELEMENT_CONTEXT,GROUP_PLAN_SCHEMA,FAILED_REGIONS,EXISTING_CANDIDATES,
    PATCH_CONTRACT,TARGET_SCHEMA,LOCAL_GRAPH,ISSUES,
    PATCH_CONTRACT_AS_NESTED_OBJECT,GRAPH,FINDINGS,TOOL_GAPS,GEOMETRY,
    PLAN,PLAN_FINGERPRINT,GROUP_REVIEW_SCHEMA,VALIDATION_ERROR},
  emphstyle=\color{annotationpromptaccent}}
\newcommand{\annotationprompt}[3]{%
  \begin{annotationpromptbox}[label={#1}]{#2}
  \lstinputlisting[style=annotationprompttext]{#3}
  \end{annotationpromptbox}}

The following reusable templates describe the annotation and refinement requests in Appendices~\ref{app:annotation-prompts} and~\ref{app:annotation-refinement}.
The grouping guidelines define a consistent annotation granularity across images; the model uses the image and element attributes to assign elements to those semantic components.
The templates also specify the structured responses needed to construct and validate the graph.
Node fields are aligned to the schema in Appendix~\ref{app:schema}.
Blue names in double braces denote inputs inserted for each request, such as the candidate inventory, current graph, or response schema.
Schema placeholders specify the required JSON fields; they do not contain reference annotations.
Shared instructions and grouping guidelines are displayed once and reused by the stages indicated below.

\paragraph{Text and non-text localization.}
The OCR model receives the image and Prompt~\ref{prompt:annotation-ocr}, shown as an English translation of the Chinese instruction.
The detector receives the image and Prompt~\ref{prompt:annotation-detector}.
Their XYXY outputs are converted to the YXYX coordinate convention before attribute annotation.

\annotationprompt{prompt:annotation-ocr}{Text localization and recognition (English translation)}{supp/prompts/01-ocr-en.txt}
\annotationprompt{prompt:annotation-detector}{Non-text candidate localization}{supp/prompts/02-detector.txt}

\paragraph{Element and scene attributes.}
Prompt~\ref{prompt:annotation-shared} precedes the attribute, scene-description, localization-recovery, crop-correction, and whole-image-review requests.
Attribute annotation receives the original image, a contextual crop, and the localized candidates; it returns attributes and uncertainties while retaining the supplied geometry and text.
Scene-description annotation receives the original image and returns the global descriptions and background.

\annotationprompt{prompt:annotation-shared}{Shared visual annotation instructions}{supp/prompts/03-shared.txt}
\annotationprompt{prompt:annotation-attributes}{Element attribute annotation}{supp/prompts/04-attributes.txt}
\annotationprompt{prompt:annotation-scene}{Scene and background descriptions}{supp/prompts/06-scene.txt}

\paragraph{Hierarchical relationship extraction.}
The grouping request receives the image and a fixed inventory of elements, excluding previous groups and parent assignments.
The inventory assigns local \texttt{E} identifiers to elements; the returned plan assigns chart and observation identifiers and declares each element's owner.
The program compiles the plan into the group types, parent links, and group attributes described in Appendix~\ref{app:annotation-prompts}.
Both grouping construction and independent grouping review insert the full policy in Prompt~\ref{prompt:annotation-grouping-policy} into \texttt{GROUPING\_POLICY}.
The construction request uses Prompt~\ref{prompt:annotation-grouping}.

\annotationprompt{prompt:annotation-grouping-policy}{Shared semantic grouping policy}{supp/prompts/07-grouping-policy.txt}
\annotationprompt{prompt:annotation-grouping}{Semantic grouping construction}{supp/prompts/07-grouping.txt}

\paragraph{Correction and review prompts.}
The templates below implement the localization-recovery, visual-correction, and grouping-review stages described in Appendix~\ref{app:annotation-refinement}.
They use the same placeholder notation as the annotation prompts.
The shared visual instructions in Prompt~\ref{prompt:annotation-shared} precede localization recovery, crop correction, and whole-image review.
Independent grouping review instead receives the full grouping policy in Prompt~\ref{prompt:annotation-grouping-policy}.

\paragraph{Localization recovery.}
Recovery requests receive a crop of the failed tool region, the region bounds, and successful candidates that must be preserved.
Text and non-text recovery use separate instructions and return only localized elements and uncertainties.
Their crop-local boxes are mapped back to full-image coordinates before subsequent annotation.

\annotationprompt{prompt:annotation-recovery-text}{Recovery of text localization}{supp/prompts/05-recovery-text.txt}
\annotationprompt{prompt:annotation-recovery-visual}{Recovery of non-text localization}{supp/prompts/05-recovery-visual.txt}

\paragraph{Local and whole-image correction.}
Prompt~\ref{prompt:annotation-patch} defines the shared patch output contract: additions or updates, removals, unresolved findings, and candidate dispositions.
It is inserted into \texttt{PATCH\_CONTRACT} for crop correction.
For whole-image review, it is inserted into \texttt{PATCH\_CONTRACT\_AS\_NESTED\_OBJECT}, with its opening phrase changed from ``Return exactly'' to ``The nested patch object contains exactly''.
Crop correction receives the original image followed by its crop, the local graph, candidates, and known issues.
Whole-image review receives the original image, complete graph, indexed findings, and tool-coverage gaps.

\annotationprompt{prompt:annotation-patch}{Shared correction-patch contract}{supp/prompts/09-patch-contract.txt}
\annotationprompt{prompt:annotation-crop}{Crop-based visual correction}{supp/prompts/10-crop-review.txt}
\annotationprompt{prompt:annotation-global}{Whole-image visual review}{supp/prompts/11-global-review.txt}

\paragraph{Grouping verification and repair.}
The independent reviewer receives the image, fixed elements, shared grouping policy, proposed plan, and its fingerprint.
Prompt~\ref{prompt:annotation-group-review} requires an accept-or-repair decision with explicit chart and observation coverage.
When a response fails structural validation, Prompt~\ref{prompt:annotation-format-repair} is appended to the corresponding request.
When grouping review rejects a plan, Prompt~\ref{prompt:annotation-group-repair} is appended to a new grouping-construction request, after which the plan is reviewed again.

\annotationprompt{prompt:annotation-group-review}{Independent grouping review}{supp/prompts/08-group-review.txt}
\annotationprompt{prompt:annotation-format-repair}{Response-format correction}{supp/prompts/12-format-repair.txt}
\annotationprompt{prompt:annotation-group-repair}{Grouping-plan correction}{supp/prompts/13-group-repair.txt}

\section{Dataset Sources, Licenses, and Extended Statistics}
\label{app:dataset-details}

\subsection{Sources and Licenses}
\label{app:dataset-sources}
\label{app:dataset-release}

\dataset{} augments real and synthetic infographic charts from ChartGalaxy~\citep{li2026chartgalaxy} with scene graph annotations.
The real charts originate from publisher websites and sharing platforms.
Table~\ref{tab:app-source-licenses} summarizes licenses and reuse conditions for selected chart sources documented by ChartGalaxy, with links to the official policies.
These include Creative Commons licenses and source-specific terms covering permitted uses, attribution, and modification.

\begin{table}[!htbp]
\centering
\caption{Selected chart sources and their licenses, following ChartGalaxy~\citep{li2026chartgalaxy}. Source names and license entries link to official policies.}
\label{tab:app-source-licenses}
\small
\renewcommand{\arraystretch}{1.15}
\begin{tabular}{@{}>{\raggedright\arraybackslash}p{0.36\linewidth}>{\raggedright\arraybackslash}p{0.34\linewidth}>{\raggedright\arraybackslash}p{0.22\linewidth}@{}}
\toprule
Name & URL & License \\
\midrule
\href{https://www.statista.com/getting-started/publishing-statista-content-daily-data---creative-commons}{Statista (Daily Data)} & \href{https://www.statista.com/}{statista.com} & \href{https://www.statista.com/getting-started/publishing-statista-content-daily-data---creative-commons}{CC BY-ND 4.0} \\
\href{https://licensing.visualcapitalist.com/terms/}{Visual Capitalist} & \href{https://www.visualcapitalist.com/}{visualcapitalist.com} & \href{https://licensing.visualcapitalist.com/terms/}{Customized} \\
\href{https://ourworldindata.org/faqs}{Our World in Data} & \href{https://ourworldindata.org/}{ourworldindata.org} & \href{https://ourworldindata.org/faqs}{CC BY} \\
\href{https://www.oecd.org/en/about/oecd-open-by-default-policy.html}{OECD} & \href{https://www.oecd.org/}{oecd.org} & \href{https://www.oecd.org/en/about/oecd-open-by-default-policy.html}{CC BY 4.0} \\
\href{https://www.ons.gov.uk/help/terms-conditions}{Office for National Statistics} & \href{https://www.ons.gov.uk/}{ons.gov.uk} & \href{https://www.ons.gov.uk/help/terms-conditions}{OGL} \\
\href{https://www.kff.org/about-us/permissions-citations-reprints/}{Kaiser Family Foundation} & \href{https://www.kff.org/}{kff.org} & \href{https://www.kff.org/about-us/permissions-citations-reprints/}{CC BY-NC-ND 4.0} \\
\href{https://informationisbeautiful.net/licensing/}{Information Is Beautiful} & \href{https://informationisbeautiful.net/}{informationisbeautiful.net} & \href{https://informationisbeautiful.net/licensing/}{Customized} \\
\href{https://chitchart.com/terms-use}{Chit Chart} & \href{https://chitchart.com/}{chitchart.com} & \href{https://chitchart.com/terms-use}{Customized} \\
\bottomrule
\end{tabular}
\end{table}

For real charts and images, we only provide their source URL, image URL, and scene graph annotations, following the original content's license and attribution requirements.

For template-based synthetic charts, ChartGalaxy documents the licenses of the input tables and visual assets~\citep{li2026chartgalaxy}.
Its reported tabular resources include UN Data (public domain), Our World in Data (CC BY), VizNet (CC BY-NC-SA), and Papers with Code (CC BY-SA).
Its visual resources include Icon645 (CC BY-NC-SA), Flaticon and Iconshock (royalty-free licenses), Heroicons (MIT), and Google Material Icons (Apache 2.0).
These licenses apply to their respective input resources, with the associated attribution and reuse conditions.

\subsection{Selection and Evaluation Splits}
\label{app:dataset-splits}

\paragraph{Subset selection.}
We target approximately 200,000 training charts with a real-to-synthetic ratio of about $1:3$, a mixture found effective in studies~\citep{rang2024scaling}.
The size target limits annotation and fine-tuning costs, while sampling source collections separately prevents the much larger layout-generated collection from determining the training mixture.
We retain real charts to cover real-world designs and use synthetic charts to expand structural and stylistic variation.
ChartGalaxy also reports gains from combining real and synthetic training data~\citep{li2026chartgalaxy}.
This yields 56,244 real and 159,951 synthetic training charts (26.0\% and 74.0\%), for a total of 216,195.
Together with the 1,000-image main test set, these form the \totalnum{} charts in \dataset{}.

\paragraph{Deduplication and split independence.}
We inherit image deduplication from ChartGalaxy, whose real-image collection uses perceptual hashing and CLIP similarity~\citep{li2026chartgalaxy}.
Subset selection preserves the original image bytes, consolidates exact duplicate image files, and excludes registered evaluation and development samples by sample identifier and image hash.
The image-to-scene-graph model is trained on the 216,195 training images without including images from the main test set.
The evaluation therefore concerns held-out chart instances and data within a shared template collection, rather than generalization to unseen templates.

\paragraph{Test sets across applications.}
The 1,000-image main test set contains 500 real charts and 500 synthetic charts, curated for source coverage and annotation quality.
It is used for both image-to-scene-graph prediction and scene graph preservation evaluation.
The QA benchmark is constructed separately from the full \dataset{} collection of \totalnum{} images and contains 1,266 questions about 1,000 charts.
The QA selection is not restricted to the main test set.
QA evaluates the benefit of annotated scene graphs during inference without additional model fine-tuning, as detailed in Appendix~\ref{app:qa-prompts}.

\subsection{Extended Annotation Statistics}
\label{app:dataset-statistics}

\paragraph{Counting conventions.}
All statistics below use the same \totalnum{} charts as Sec.~\ref{sec:statistics}.
We distinguish the 3,088,632 annotated group nodes from the implicit root of each image.
Including these roots yields \groupnum{} groups, matching the main-text average of \avggroupnum{} groups per chart.
The root is excluded from the seven group-type counts in Table~\ref{tab:app-node-distribution}.
Each stored element or group has one parent, so the number of hierarchical edges is the number of elements plus the number of non-root groups: $15{,}591{,}656 + 3{,}088{,}632 = 18{,}680{,}288$.
The depth of a scene graph is the number of nodes on its longest root-to-element path, counting both the root and the terminal element.
A graph whose elements are all direct children of the root therefore has depth two.

\begin{table}[!htbp]
\centering
\caption{Element and group-type distributions. Percentages use all \elementnum{} elements or all 3,088,632 non-root groups as the denominator, respectively.}
\label{tab:app-node-distribution}
\small
\begin{tabular}{@{}llrr@{}}
\toprule
Node kind & Type & Count & Percentage \\
\midrule
Element & \texttt{text} & 6,044,031 & 38.76\% \\
 & \texttt{image} & 1,107,791 & 7.11\% \\
 & \texttt{shape} & 8,439,834 & 54.13\% \\
\midrule
Group & \texttt{chart} & 221,006 & 7.16\% \\
 & \texttt{axis} & 72,699 & 2.35\% \\
 & \texttt{legend} & 88,976 & 2.88\% \\
 & \texttt{legend\_item} & 166,058 & 5.38\% \\
 & \texttt{data\_item} & 2,074,082 & 67.15\% \\
 & \texttt{series} & 73,181 & 2.37\% \\
 & \texttt{panel} & 392,630 & 12.71\% \\
\bottomrule
\end{tabular}
\end{table}

\noindent\begin{minipage}{\linewidth}
\paragraph{Graph size and hierarchy.}
Table~\ref{tab:app-graph-distribution} summarizes the per-chart distributions without truncating large graphs.
The median chart contains 54 elements and 10 non-root groups, while the 99th percentiles are 336 and 57, respectively.
Nested groups occur in 211,736 charts (97.49\%): at least one non-root group directly contains another group.
The median hierarchy depth is four, with a maximum of eight.
\end{minipage}

\begin{table}[!htbp]
\centering
\caption{Per-chart distributions over all \totalnum{} charts. Total nodes include one implicit root per chart; non-root groups exclude it. Quantiles use linear interpolation of the ordered observations.}
\label{tab:app-graph-distribution}
\small
\setlength{\tabcolsep}{4pt}
\begin{tabular}{@{}lrrrrrrrr@{}}
\toprule
Quantity & Min & P25 & Median & P75 & P90 & P99 & Max & Mean \\
\midrule
Elements & 1 & 35 & 54 & 85 & 121 & 336 & 3,682 & 71.79 \\
Non-root groups & 0 & 7 & 10 & 18 & 31 & 57 & 234 & 14.22 \\
Total nodes & 2 & 44 & 67 & 108 & 151 & 344 & 3,725 & 87.01 \\
Hierarchy depth & 2 & 4 & 4 & 5 & 6 & 7 & 8 & 4.54 \\
\bottomrule
\end{tabular}
\end{table}

\begin{figure}[!htbp]
\centering
\includegraphics[width=\linewidth]{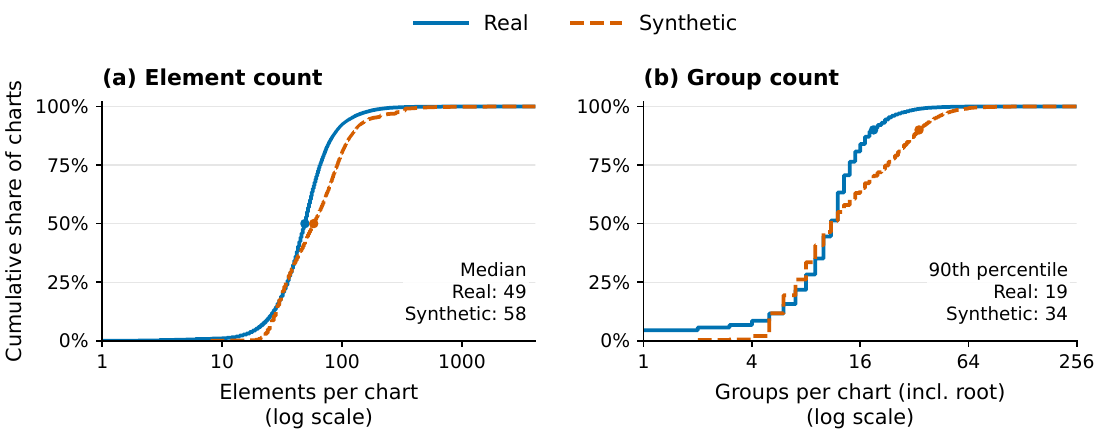}
\caption{Empirical cumulative distributions of (a) element counts and (b) group counts for real and synthetic infographic charts. Both count axes are logarithmic and include all charts. Markers identify the median element count and the 90th-percentile group count. Each distribution is normalized within its real or synthetic subset. Group counts include the implicit root, matching the main-text convention.}
\label{fig:app-structure-distributions}
\end{figure}

Fig.~\ref{fig:app-structure-distributions} compares the annotated structure of real and synthetic charts.
Synthetic charts have a higher median element count (58 versus 49).
Both subsets have a median of 11 groups when the root is included, but their 90th-percentile group counts differ: 34 for synthetic charts and 19 for real charts.
Thus, the same median group count masks a difference in the frequency of charts with many groups.

\section{Image-to-Scene-Graph Prediction}
\label{app:i2sg}

\subsection{Training Configurations}
\label{app:i2sg-training}

We build our image-to-scene-graph model through supervised fine-tuning of Qwen3.5-4B~\citep{qwen2026qwen35} on the \dataset{} training set.
We perform full-parameter fine-tuning for $100{,}000$ steps. 
Training samples are packed into sequences of up to $14{,}336$ tokens, with a global batch size of $32$ packed sequences. 
We use the Adam optimizer with $\beta_1=0.9$, $\beta_2=0.95$, $\epsilon=10^{-8}$, weight decay $0.1$, and gradient-norm clipping at $1.0$. 
The learning rate warms up from zero to $5\times10^{-5}$ over the first $500$ steps, followed by cosine decay to zero over the remainder of training.
The target serializes node attributes and parent assignments, excluding spatial relationships that are derived from bounding boxes.
The reported model uses the checkpoint at step $100{,}000$.
At inference, it uses vLLM 0.20.2, a context limit of $32{,}768$ tokens, a fixed output limit of $20{,}000$ tokens, temperature 0, and top-$p$ of 1.

\subsection{Baseline Settings}
\label{app:i2sg-baselines}
All models are evaluated on the same 1,000 test images and reference graphs.
Inference receives the original image and task specification, without reference annotations or source data tables.
The common task predicts node attributes and hierarchical relationships; spatial relationships are computed afterward.
The recorded runs use model-specific serialization and decoding settings, summarized in Table~\ref{tab:app-i2sg-settings}.
Thus, the comparison evaluates the specified inference pipelines, whose call counts and token budgets differ.

\begin{table}[!htbp]
\centering
\caption{Inference settings for image-to-scene-graph prediction. Context and output limits are in tokens; an API dash denotes a context limit not explicitly set by the client. Open-model revision prefixes identify the archived weight revisions; full identifiers are retained in the accompanying inference configuration files.}
\label{tab:app-i2sg-settings}
\small
\setlength{\tabcolsep}{4pt}
\begin{tabular}{@{}lrrl@{}}
\toprule
Model (revision prefix) & Context & Output & Decoding \\
\midrule
Molmo2-8B (\texttt{e28fa28}) & 36,864 & 24,576 & Sampling; inventory first \\
InternVL3.5-8B (\texttt{9bb6a56}) & 32,768 & 24,576 & Temperature 0 \\
Gemma 4-12B-IT (\texttt{707f0a3}) & 65,536 & 24,576 & Temperature 0 \\
Qwen3.5-4B (\texttt{851bf6e}) & 32,768 & 24,576 & Sampling; thinking off \\
Qwen3.8-27B (\texttt{1d4bf0f}) & 65,536 & 24,576 & Temperature 0; thinking off \\
Qwen3.8-Flash-Next-FP8 (\texttt{236dfdf}) & 65,536 & 24,576 & Temperature 0; thinking off \\
GLM-5.3-Flash (\texttt{eb9eb20}) & 65,536 & 24,576 & Temperature 0; high reasoning \\
\midrule
Gemini 3.8 Flash & -- & 24,576 & LOW thinking \\
GPT-5.6 Sol & -- & 24,576 & Medium reasoning \\
GPT-6 Astra & -- & 24,576 & Medium reasoning \\
\bottomrule
\end{tabular}
\end{table}

\paragraph{Decoding and format adaptation.}
The two GPT models use structured JSON output and one prediction request per image, without format-feedback generation.
Their API identifiers are \texttt{gpt-5.6-sol} and \texttt{gpt-6-astra}.
Gemini uses \texttt{gemini-3.8-flash}, a JSON response schema, and up to three generation attempts with validation feedback, stopping at the first valid graph.
Open models use a compact JSON grammar and up to three generation attempts under the same stopping rule.
Molmo2 instead uses one unconstrained visual-inventory request, capped at 8,192 tokens, followed by at most two graph requests; up to 10,000 characters of its inventory are included in the graph prompt.
For Qwen3.5-4B and Molmo2, sampling uses temperature 0.7, top-$p$ 0.8, top-$k$ 20, minimum-$p$ 0, presence penalty 1.5, repetition penalty 1, and seed 20260916.
Recorded operational retries address request failures and do not select predictions by evaluation score.

GPT predicts native $[y_0,x_0,y_1,x_1]$ boxes and hexadecimal colors.
The other baseline interfaces use $[x_0,y_0,x_1,y_1]$ boxes and integer RGBA tuples, except that Molmo2 names the four coordinate fields explicitly.
These forms are converted deterministically to the common coordinate and color representation before scoring.
Reversed endpoint pairs are sorted within their own axes; unambiguous label renaming updates all associated references.
These operations do not use reference graphs or estimate missing coordinates.

\paragraph{Invalid and partial outputs.}
Every test image remains in the evaluation.
An unrecovered failure contributes zero node and spatial F1, and its reference parent edges remain in the hierarchical denominator.
The recorded InternVL3.5 and Molmo2 recovery retains individually evaluable nodes from partial outputs while preserving rejected records in the node and parent denominators.
Nodes with usable boxes can contribute predicted spatial relations even when their other fields prevent matching; records without usable boxes create no predicted near edges, while missed reference near edges remain false negatives and node and parent penalties are retained.
For our model, the fixed postprocessing also retains parseable prefixes of truncated outputs, removes identical duplicate keys, and resolves unambiguous serialization inconsistencies.
An incomplete terminal node may be quarantined; missing visual objects are not generated during this recovery.
These are scores of the recorded prediction and postprocessing pipelines, rather than raw JSON-validity rates.

\subsection{Evaluation Metrics}
\label{app:metrics}

\subsubsection{Node Matching}
\label{app:node-matching}
The node-level columns in Table~\ref{tab:image2sg} evaluate both element and group nodes.
Let $V_i$ and $\widehat V_i$ be the reference and predicted node sets for image $i$, excluding the virtual \texttt{ROOT}.
Global style and composition descriptions are not scored as nodes.
Only nodes of the same kind, element or group, are eligible to match.
Their concrete types need not agree: type agreement is a scored attribute rather than a hard constraint.
Node labels and parent assignments do not enter the matching objective, and there is no IoU threshold.

For an eligible reference--prediction pair $(v,\hat v)$, the applicable attribute set $\mathcal A(v)$ always contains type, box, and description.
It additionally contains text when the reference has nonempty normalized text, and color when its palette is nonempty.
Applicability is determined by the reference; omitting a required predicted attribute does not remove it from the score.
All five attributes have nominal weight 0.2, renormalized over $\mathcal A(v)$.
Thus, the pair score is their arithmetic mean:
\begin{equation}
\widetilde s(v,\hat v)=\frac{1}{|\mathcal A(v)|}
\sum_{a\in\mathcal A(v)}\widetilde s_a(v,\hat v).
\end{equation}
We apply maximum-weight Hungarian assignment using $\widetilde s$, assigning zero weight to ineligible pairs and discarding zero-weight assignments.
Denote the retained one-to-one matches by $M_i$ and the induced prediction-to-reference map by $\pi_i$.
We additionally set $\pi_i(\texttt{ROOT})=\texttt{ROOT}$ for hierarchical evaluation.

To control the cost of description judging, we use a lightweight lexical proxy during assignment and apply the description judge only to the matched node pairs.
After case folding and whitespace collapse, the proxy is the maximum of RapidFuzz string-ratio and token-set-ratio similarities, divided by 100.
After fixing $M_i$, we replace this proxy with the description-judge score in Appendix~\ref{app:attribute-similarities} and recompute the pair mean $s(v,\hat v)$.
We do not rerun assignment after judging.
All reported attributes and relationship scores use this same fixed matching.

\subsubsection{Attribute Similarities}
\label{app:attribute-similarities}
\paragraph{Type.}
Type similarity is 1 for equal concrete types and 0 otherwise.
For elements, the types are text, image, and shape; for groups, they are the semantic group types.

\paragraph{Bounding box.}
For boxes $B$ and $\widehat B$, let $I$, $U$, and $C$ be the areas of their intersection, union, and smallest enclosing axis-aligned rectangle.
We normalize GIoU~\citep{Rezatofighi2019GIoU} to $[0,1]$:
\begin{equation}
s_{\mathrm{box}}=\frac{1}{2}\left(1+\frac{I}{U}-\frac{C-U}{C}\right).
\end{equation}
The normalized score is used both during assignment and in the reported Box column.
Scored boxes have positive area and coordinates within the normalized image canvas.

\paragraph{Text.}
We apply Unicode NFKC normalization, replace the Unicode minus sign with an ASCII hyphen, and collapse whitespace without changing letter case.
For resulting strings $t$ and $\hat t$, the base similarity is
\begin{equation}
s_{\mathrm{text}}^{\mathrm{base}}=1-\frac{\operatorname{Lev}(t,\hat t)}{\max(|t|,|\hat t|)}.
\end{equation}
A missing or empty prediction receives zero when reference text is nonempty.
To prevent long strings from masking a wrong value, we cap the score at 0.5 when the ordered lists of numeric values, adjacent units, or currency signs differ.
Numeric comparison normalizes equivalent decimal/scientific spellings and unambiguous three-digit comma grouping; unit spelling and case are retained.
The rule does not infer locale or convert between units.

\paragraph{Color palette.}
We canonicalize hexadecimal RGB/RGBA colors, expand three-digit RGB, set omitted alpha to 1, remove duplicates, and identify all fully transparent colors with one transparent entry.
For colors $c$ and $\hat c$ with alpha values $\alpha$ and $\hat\alpha$, define
\begin{equation}
k(c,\hat c)=\max\left(0,1-\frac{\Delta E_{00}(c,\hat c)}{100}\right)
(1-|\alpha-\hat\alpha|),
\end{equation}
where $\Delta E_{00}$ is CIEDE2000~\citep{luo2001development} on the RGB colors converted to CIELAB.
When both colors are fully transparent, $k=1$ regardless of RGB.
For reference palette $\mathcal C$ and predicted palette $\widehat{\mathcal C}$, we measure bidirectional coverage:
\begin{equation}
P_c=\frac{\sum_{\hat c\in\widehat{\mathcal C}}\max_{c\in\mathcal C}k(c,\hat c)}{|\widehat{\mathcal C}|},
\qquad
R_c=\frac{\sum_{c\in\mathcal C}\max_{\hat c\in\widehat{\mathcal C}}k(c,\hat c)}{|\mathcal C|},
\qquad
s_{\mathrm{color}}=\frac{2P_cR_c}{P_c+R_c}.
\end{equation}
Colors may provide coverage to more than one counterpart; this is not one-to-one palette assignment.
An empty predicted palette or zero combined coverage receives zero.
An empty reference palette makes this attribute inapplicable.

\paragraph{Description.}
For each matched pair, GPT-5.6 Luna compares the predicted and reference description strings.
It receives no image and measures preservation of reference meaning, rather than independently verifying visual truth.
Each request contains one comparison, uses low reasoning effort and a 4,096-token output cap, and requires a strict JSON result with an integer score and rationale.
The returned score $q\in\{0,1,2,3,4,5\}$ gives $s_{\mathrm{description}}=q/5$.
Harmless paraphrases retain credit; omissions, unsupported additions, and contradictions reduce the score.

\subsubsection{Node and Edge Scores}
\label{app:graph-scores}
\paragraph{Overall node score and attribute columns.}
Let $S_i=\sum_{(v,\hat v)\in M_i}s(v,\hat v)$.
The similarity-weighted node F1 for image $i$ is
\begin{equation}
P_i^{\mathrm{node}}=\frac{S_i}{|\widehat V_i|},\qquad
R_i^{\mathrm{node}}=\frac{S_i}{|V_i|},\qquad
F_i^{\mathrm{node}}=\frac{2S_i}{|V_i|+|\widehat V_i|}.
\end{equation}
Unmatched nodes receive no credit but remain in the corresponding denominators.
The direct F1 expression also gives zero for an empty prediction when the reference is nonempty.
The Overall column averages $F_i^{\mathrm{node}}$ over all $N=1{,}000$ images.
Each separate attribute column instead pools the sum and count of its applicable matched-node scores across images.
These conditional attribute means describe matched-node quality; the Overall column additionally penalizes omissions and extra nodes.

\paragraph{Hierarchical relationships.}
Each non-root node contributes one direct parent edge, including an edge to \texttt{ROOT} when appropriate.
A predicted edge is correct when its child is matched and its parent maps to the reference child's parent:
\begin{equation}
C_i^h=\sum_{(v,\hat v)\in M_i}
\mathbf{1}\!\left[\pi_i(\operatorname{parent}(\hat v))=\operatorname{parent}(v)\right].
\end{equation}
Unknown, unmatched, or invalid parents receive no credit.
We score direct parents, without substituting ancestor agreement.
\begin{equation}
F^h=\frac{2\sum_i C_i^h}{\sum_i(|V_i|+|\widehat V_i|)}.
\end{equation}

\paragraph{Spatial relationships.}
We derive geometry for all selected pairs of non-root nodes on each side, independently of the sparse serialized relationship selection in Appendix~\ref{app:spatial-relations}.
One node pair and one relationship type form one scoring unit.
Direction, alignment, and overlap use the bounding-box predicates in Appendix~\ref{app:spatial-relations}, with alignment tolerance 5.
A diagonal direction such as \texttt{top-left} is one directional relationship, and all satisfied alignment bases form one alignment relationship.
Only nonempty direction/alignment and positive-area overlap are counted; agreement on absence is not credited.
A categorical relation receives one unit of credit only if both endpoints match and the direction or complete alignment set, or the overlap predicate, agrees under the mapped endpoint order.

The proximity component scores positive Boolean near relationships using the predicate in Appendix~\ref{app:spatial-relations}, with the actual image width and height on each side.
Let $E_i^{\mathrm{near}}$ and $\widehat E_i^{\mathrm{near}}$ be the reference and predicted near-pair sets over all unordered non-root node pairs.
A predicted near pair receives one unit of credit only when both endpoints are matched and their mapped reference pair is also near:
\begin{equation}
C_{i,\mathrm{near}}^s=\sum_{\{\hat u,\hat v\}\in\widehat E_i^{\mathrm{near}}}
\mathbf{1}\!\left[\{\pi_i(\hat u),\pi_i(\hat v)\}\in E_i^{\mathrm{near}}\right],
\end{equation}
where an unmatched endpoint makes the indicator zero.
The proximity reference and prediction counts are $|E_i^{\mathrm{near}}|$ and $|\widehat E_i^{\mathrm{near}}|$; their F1 is $2C_{i,\mathrm{near}}^s/(|E_i^{\mathrm{near}}|+|\widehat E_i^{\mathrm{near}}|)$ when the denominator is positive.
If both sets are empty, the proximity component is undefined and omitted from its component mean; its zero counts contribute nothing to the pooled spatial score below.
True negatives receive no credit, and continuous distance similarity is not scored.
We recompute the predicate from boxes, so changing which relations are stored does not change the reported spatial score.
All reported Spatial scores for image-to-scene-graph prediction and scene graph preservation use this Boolean rule, while keeping the recorded model outputs, node matches, and other spatial components fixed.

Let $C_{i,t}^s$, $R_{i,t}^s$, and $P_{i,t}^s$ be the credit, reference count, and prediction count for spatial type $t$.
We pool the four types within each image, then average images equally:
\begin{equation}
F_i^s=\frac{2\sum_t C_{i,t}^s}{\sum_t(R_{i,t}^s+P_{i,t}^s)},\qquad
F^s=\frac{1}{N}\sum_i F_i^s.
\end{equation}
For partial outputs, missing or invalid boxes create no predicted near edges; reference near edges without a corresponding correct prediction remain false negatives.
Spatial counts use available geometry, while the node and parent penalties above remain unchanged.
Every reference image in this cohort has spatial relationships, so the per-image denominator is nonzero.

\subsection{Further Analysis}
\label{app:image2sg-complexity}
We analyze how extraction performance varies with the number of reference nodes, counting both elements and groups and excluding \texttt{ROOT}.
The 1,000 test images are divided into five bins: $n<50$, $50\leq n<75$, $75\leq n<100$, $100\leq n\leq125$, and $n>125$, containing 191, 218, 193, 178, and 220 images, respectively.
All three metrics are averaged equally over images within each bin.
Predictions, node assignments, and per-image scores are unchanged; all models use the same images, including failed outputs.

\begin{figure}[!htbp]
\centering
\includegraphics[width=\linewidth]{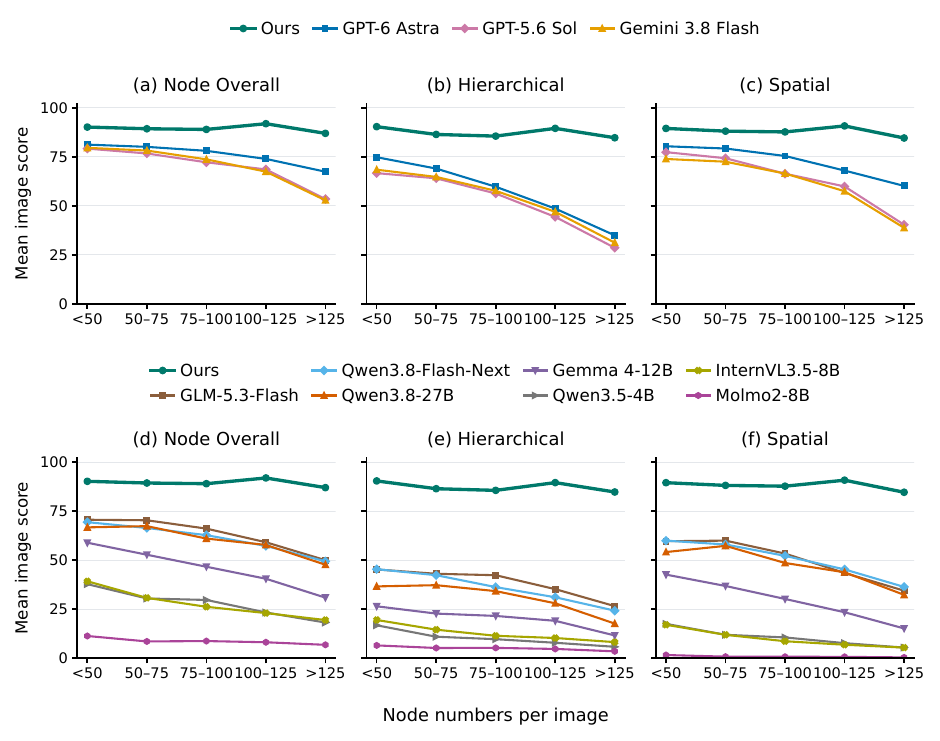}
\caption{Performance by reference node count.
The top row compares our model with the three proprietary baselines; the bottom row includes all seven open-source baselines.
All panels use equal image weights within each bin.}
\label{fig:image2sg-complexity-supp}
\end{figure}

Figure~\ref{fig:image2sg-complexity-supp} reports all eleven models.
Our model has the highest Overall, Hierarchical, and Spatial scores in each of the five bins.
The leading baselines generally obtain lower scores on images with more nodes, while our model has smaller differences between the first and last bins.
For Hierarchical, GPT-6 Astra decreases from 74.8 in the $n<50$ bin to 35.0 in the $n>125$ bin, whereas ours changes from 90.4 to 84.7.
The corresponding gap increases from 15.6 to 49.7 points.
Overall and Spatial show a similar increase in the first-to-last-bin gap.
Our scores do not decrease monotonically: they rise in the 100--125-node bin.
Because the bins contain different mixtures of image sources, these curves characterize observed performance across graph sizes rather than isolate a causal effect of node count.

\definecolor{qaaccent}{HTML}{31556F}
\definecolor{qalight}{HTML}{F4F7FA}
\definecolor{qarule}{HTML}{BCCBD5}
\newtcolorbox[auto counter,number within=section]{qapromptbox}[2][]{
  enhanced,breakable,colback=qalight,colframe=qarule,
  colbacktitle=qaaccent!10!white,coltitle=black,
  fonttitle=\small\bfseries,fontupper=\small,
  boxrule=0.45pt,arc=1mm,left=8pt,right=8pt,top=7pt,bottom=7pt,
  before skip=9pt,after skip=9pt,
  title={Prompt~\thetcbcounter. #2},
  title after break={Prompt~\thetcbcounter. #2 (continued)},#1}
\newcommand{\qavariable}[1]{{\color{qaaccent}\ttfamily\{#1\}}}
\lstdefinestyle{qajson}{basicstyle=\ttfamily\footnotesize,
  columns=fullflexible,keepspaces=true,showstringspaces=false,
  breaklines=true,aboveskip=5pt,belowskip=5pt,
  backgroundcolor=\color{white},frame=single,rulecolor=\color{qarule},
  framerule=0.3pt,xleftmargin=3pt,xrightmargin=3pt}
\newtcolorbox{qaexamplebox}[1]{enhanced,colback=white,colframe=qarule,
  colbacktitle=qalight,coltitle=black,fonttitle=\small\bfseries,fontupper=\small,
  boxrule=0.4pt,arc=0.6mm,left=6pt,right=6pt,top=6pt,bottom=6pt,
  valign=top,title={#1}}

\section{Scene-Graph-Augmented Question Answering}
\label{app:qa}

\subsection{Example Question-Answer Pairs}
\label{app:qa-examples}

\begin{figure}[!htbp]
\centering
\begin{qaexamplebox}{(a) Within-group correspondences}
\noindent\begin{minipage}[c]{0.245\linewidth}
\centering
{\scriptsize\bfseries Full chart}\par\smallskip
\begin{tikzpicture}
\node[anchor=south west,inner sep=0] (overview) at (0,0)
{\includegraphics[width=0.98\linewidth,height=0.275\textheight,keepaspectratio]{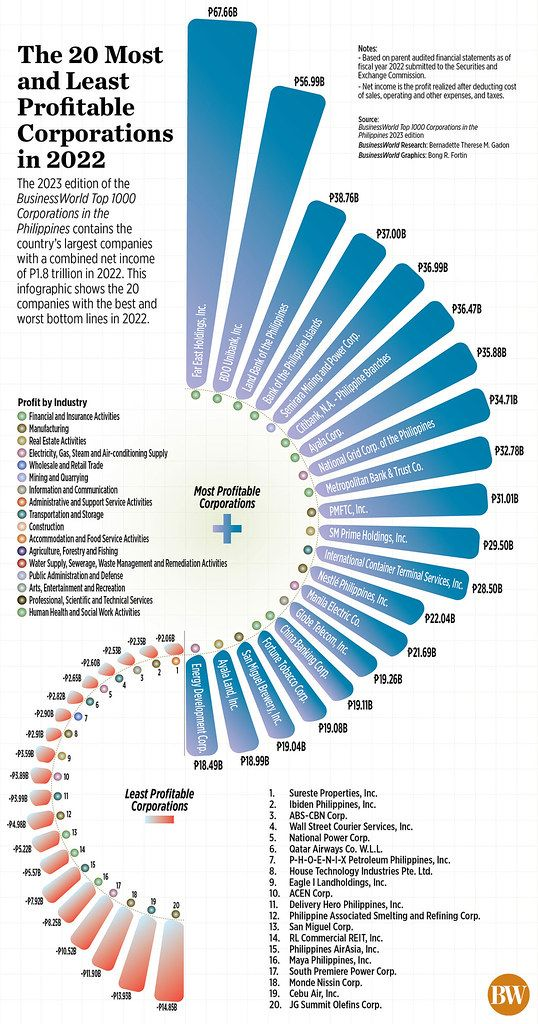}};
\begin{scope}[x={(overview.south east)},y={(overview.north west)}]
\draw[qaaccent,line width=0.8pt] (0.390335,0.492188) rectangle (1.000000,0.853516);
\end{scope}
\end{tikzpicture}
\end{minipage}\hfill
\begin{minipage}[c]{0.035\linewidth}
\centering\textcolor{qaaccent}{$\rightarrow$}
\end{minipage}\hfill
\begin{minipage}[c]{0.685\linewidth}
\centering
{\scriptsize\bfseries Enlarged region}\par\smallskip
\includegraphics[trim=210bp 504bp 0bp 150bp,clip,width=\linewidth,height=0.275\textheight,keepaspectratio]{figs/qa/qa_profit.png}
\end{minipage}
\par\medskip
\textbf{Question.} Which corporation has a profit of PHP\,38.76B?\par\smallskip
\textbf{Answer.} Land Bank of the Philippines
\end{qaexamplebox}
\begin{qaexamplebox}{(b) Across-group reasoning}
\noindent\begin{minipage}[c]{0.245\linewidth}
\centering
{\scriptsize\bfseries Full chart}\par\smallskip
\begin{tikzpicture}
\node[anchor=south west,inner sep=0] (overview) at (0,0)
{\includegraphics[width=0.98\linewidth,height=0.275\textheight,keepaspectratio]{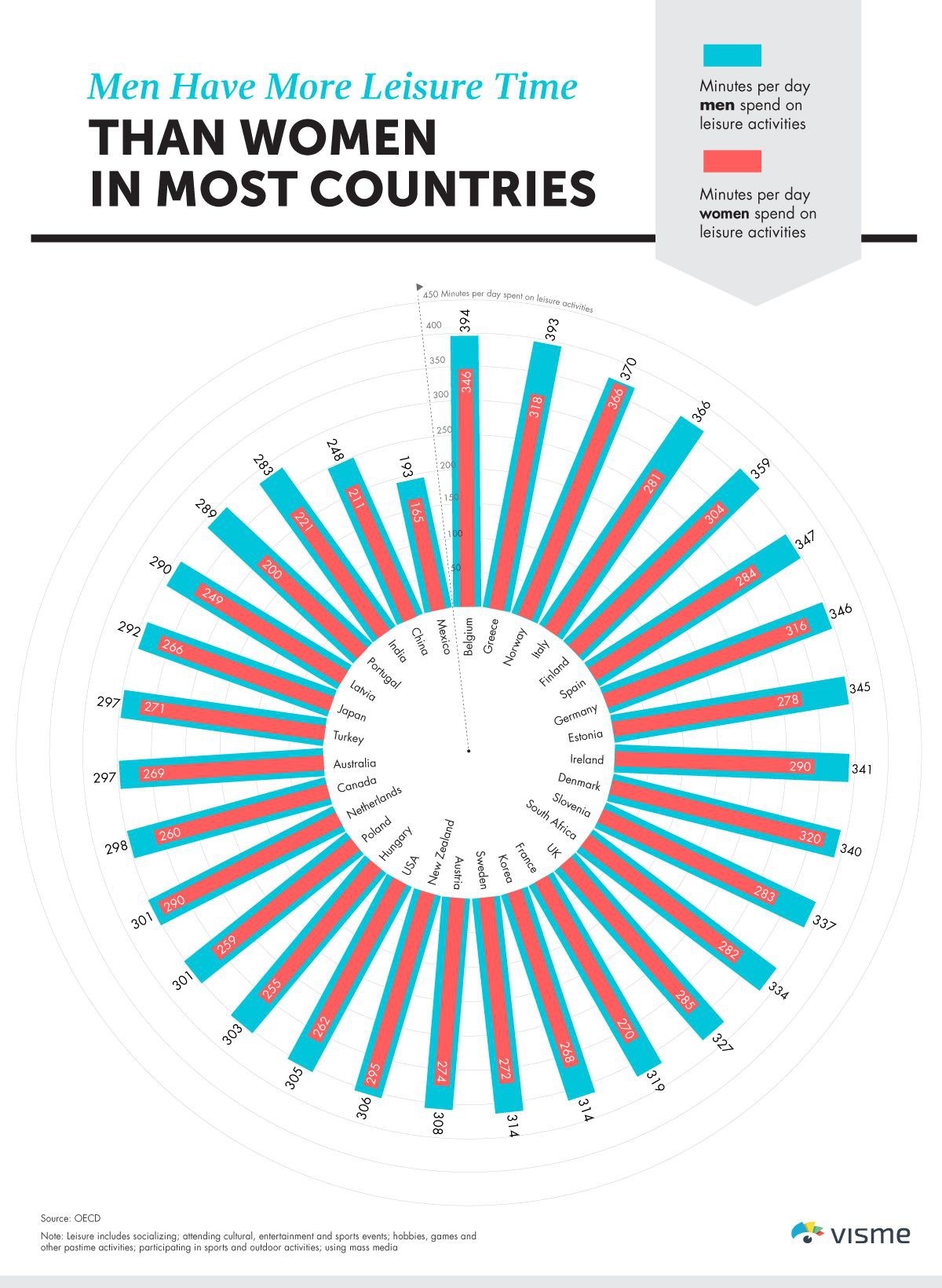}};
\begin{scope}[x={(overview.south east)},y={(overview.north west)}]
\draw[qaaccent,line width=0.8pt] (0.141667,0.359756) rectangle (0.841667,0.713415);
\end{scope}
\end{tikzpicture}
\end{minipage}\hfill
\begin{minipage}[c]{0.035\linewidth}
\centering\textcolor{qaaccent}{$\rightarrow$}
\end{minipage}\hfill
\begin{minipage}[c]{0.685\linewidth}
\centering
{\scriptsize\bfseries Enlarged region}\par\smallskip
\includegraphics[trim=170bp 590bp 190bp 470bp,clip,width=\linewidth,height=0.275\textheight,keepaspectratio]{figs/qa/qa_leisure.jpg}
\par{\scriptsize Blue: men; red: women. Values are minutes per day.}
\end{minipage}
\par\medskip
\textbf{Question.} Which country has the larger gap between men's and women's daily leisure time, Portugal or Italy?\par\smallskip
\textbf{Answer.} Portugal
\end{qaexamplebox}
\caption{QA examples of within-group correspondences and across-group reasoning. Boxes on the full charts mark the regions enlarged on the right.}
\label{fig:qa-example-images}
\end{figure}

\begin{figure}[!htbp]
\centering
\begin{qaexamplebox}{(c) Connected elements}
\noindent\begin{minipage}[c]{0.245\linewidth}
\centering
{\scriptsize\bfseries Full chart}\par\smallskip
\begin{tikzpicture}
\node[anchor=south west,inner sep=0] (overview) at (0,0)
{\includegraphics[width=0.98\linewidth,height=0.275\textheight,keepaspectratio]{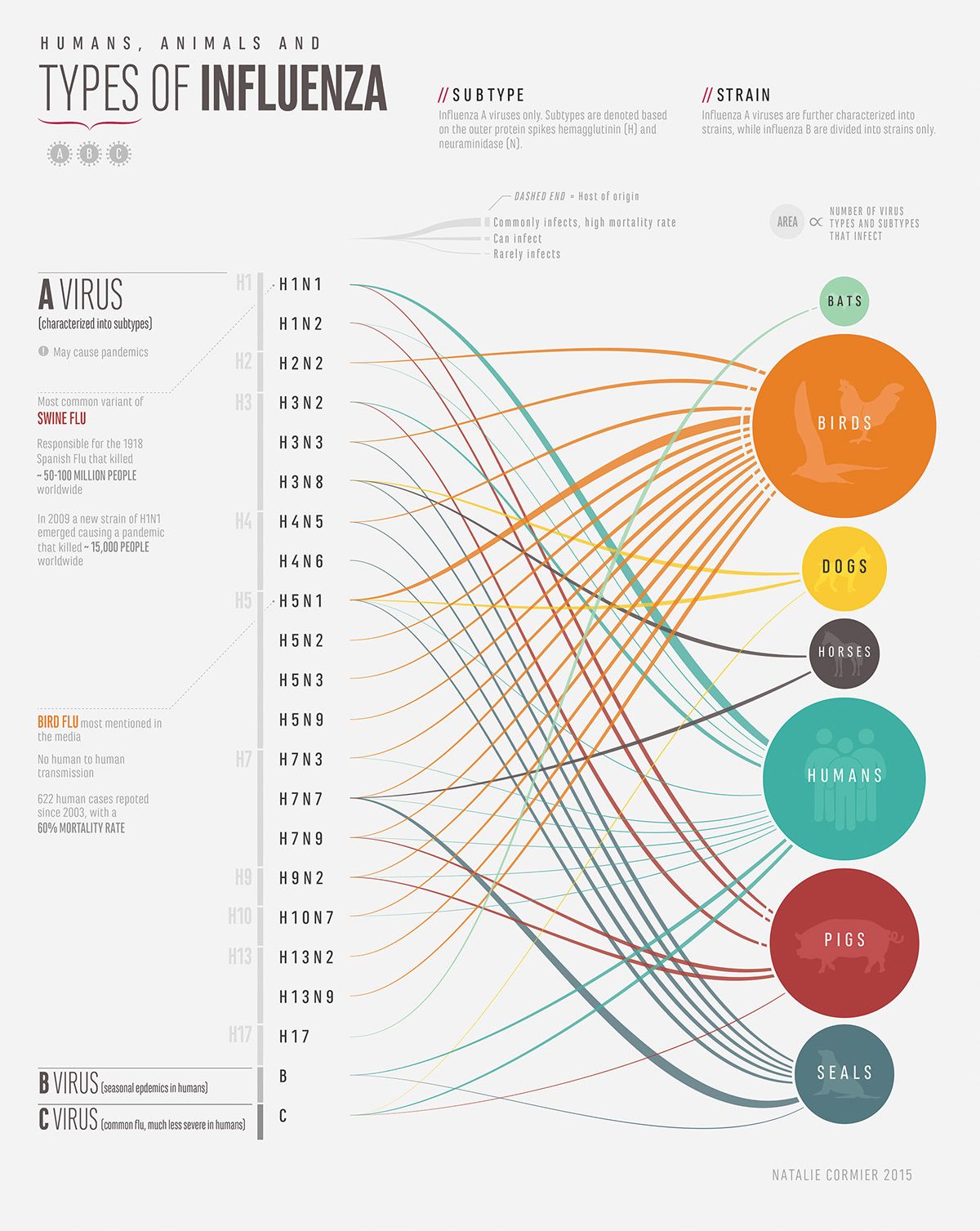}};
\begin{scope}[x={(overview.south east)},y={(overview.north west)}]
\draw[qaaccent,line width=0.8pt] (0.258333,0.072993) rectangle (0.954167,0.784340);
\end{scope}
\end{tikzpicture}
\end{minipage}\hfill
\begin{minipage}[c]{0.035\linewidth}
\centering\textcolor{qaaccent}{$\rightarrow$}
\end{minipage}\hfill
\begin{minipage}[c]{0.685\linewidth}
\centering
{\scriptsize\bfseries Enlarged region}\par\smallskip
\includegraphics[trim=310bp 110bp 55bp 325bp,clip,width=\linewidth,height=0.275\textheight,keepaspectratio]{figs/qa/qa_connectors.jpg}
\end{minipage}
\par\medskip
\textbf{Question.} Which hosts are connected to H4N5 or H4N6?\par\smallskip
\textbf{Answer.} Birds, Seals
\end{qaexamplebox}
\begin{qaexamplebox}{(d) Adjacent elements}
\noindent\begin{minipage}[c]{0.245\linewidth}
\centering
{\scriptsize\bfseries Full chart}\par\smallskip
\begin{tikzpicture}
\node[anchor=south west,inner sep=0] (overview) at (0,0)
{\includegraphics[width=0.98\linewidth,height=0.275\textheight,keepaspectratio]{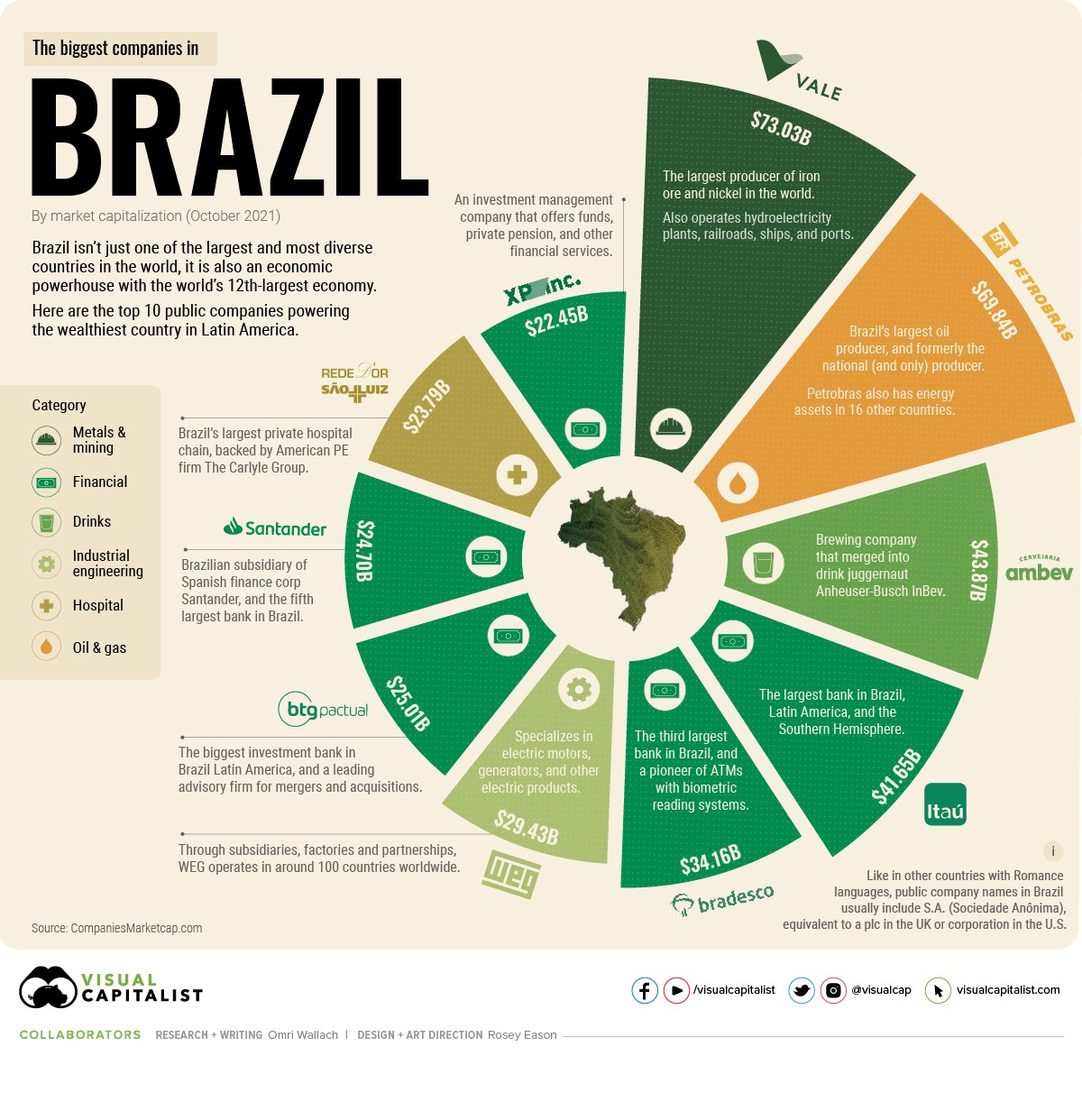}};
\begin{scope}[x={(overview.south east)},y={(overview.north west)}]
\draw[qaaccent,line width=0.8pt] (0.187500,0.380032) rectangle (0.616667,0.778583);
\end{scope}
\end{tikzpicture}
\end{minipage}\hfill
\begin{minipage}[c]{0.035\linewidth}
\centering\textcolor{qaaccent}{$\rightarrow$}
\end{minipage}\hfill
\begin{minipage}[c]{0.685\linewidth}
\centering
{\scriptsize\bfseries Enlarged region}\par\smallskip
\includegraphics[trim=225bp 472bp 460bp 275bp,clip,width=\linewidth,height=0.275\textheight,keepaspectratio]{figs/qa/qa_neighbors.jpg}
\end{minipage}
\par\medskip
\textbf{Question.} Which two companies immediately border the wedge labeled \$23.79B?\par\smallskip
\textbf{Answer.} Santander, XP Inc.
\end{qaexamplebox}
\caption{QA examples of connections and spatial adjacency. Boxes on the full charts mark the regions enlarged on the right.}
\label{fig:qa-example-images-relations}
\end{figure}

Figures~\ref{fig:qa-example-images} and~\ref{fig:qa-example-images-relations} present examples from real infographic charts.
Each row pairs the full chart and an enlarged region with its question and reference answer.

\clearpage

\subsection{Details of the Scoring Rules}
\label{app:qa-metric}

\paragraph{Answer extraction.}
Direct responses are read as the answer, while guided responses are scored using only \texttt{final\_answer}.
The parser also accepts an unambiguous final-answer field or section surrounded by imperfect formatting, without consulting the reference answer.
Empty or unrecoverable answers count as incorrect, including outputs truncated before the final answer.
Accuracy gives equal weight to every question.

\paragraph{Answer-type-specific matching.}
We use the InfoChartQA evaluator~\citep{lin2025infochartqa} with the stored answer type.
Text matching is case-insensitive: the released checker accepts exact or contained reference strings, otherwise requiring a \texttt{SequenceMatcher} ratio of at least 0.8.
Numerical answers use the evaluator's parsing and supported unit, percentage, and magnitude conversions.
A converted reference--prediction pair $(v,\hat v)$ is accepted when
\[
\frac{|v-\hat v|}{0.01+|v|}<\epsilon,
\]
with $\epsilon=0.05$ for relaxed numerical questions and $\epsilon=0.001$ for the stricter branch.
The mixed value/text type selects a branch from the reference's format.
Lists use ordered or unordered matching; ordered answers must preserve both order and length.
Fixed-form answers, such as yes/no responses, use the corresponding evaluator branch.

\paragraph{Equivalent formats.}
Following InfoChartQA~\citep{lin2025infochartqa}, equivalent answer representations are normalized before comparison, covering numerical expressions (e.g., ``1K'' and ``1,000''), date formats (e.g., ``January 1, 2024'' and ``2024-01-01''), list representations (e.g., ``A, B'' and ``A and B''), and predefined name aliases (e.g., ``USA'' and ``United States'').

\subsection{Prompting Settings}
\label{app:qa-prompts}

\paragraph{Evaluation protocol.}
All QA models are evaluated without additional fine-tuning for this study.
All prompting conditions use the same QA charts and questions.
We use low reasoning effort for each model and across the three prompting conditions.
The comparison measures the benefit of supplying annotated scene graphs at inference time.

\paragraph{Scene graph input.}
The scene graphs provided in the QA experiments are produced by our dataset annotation pipeline rather than predicted by the fine-tuned image-to-scene-graph model.
Before constructing the QA input, we convert the normalized $[y_0,x_0,y_1,x_1]$ boxes in Appendix~\ref{app:node-attributes} to $[x_{\min},y_{\min},x_{\max},y_{\max}]$ in original-image pixels.
For an image of width $W$ and height $H$, this conversion reorders the coordinates and scales the horizontal and vertical components by $W/1000$ and $H/1000$, respectively.
The input is serialized as JSON containing nodes and spatial relations.
Each node stores a concise label, type, bounding box, parent reference, and available text; spatial relations refer to the same labels.
A null parent denotes a root.
Labels describe element roles and associated entities using the descriptions and grouping context.

\paragraph{Prompt templates.}
Boxes~\ref{box:qa-direct} and~\ref{box:qa-guided} reproduce the evaluated system instructions.
Box~\ref{box:qa-user} shows the user-input template; blue placeholders are replaced for each question.
The reference answer is withheld.

\begin{qapromptbox}[label=box:qa-direct]{System instruction: Direct prompting}
Answer the chart question from the image and follow the requested answer format exactly.
\end{qapromptbox}

\begin{qapromptbox}[label=box:qa-user]{User input: Shared template}
\textbf{Text}\par\smallskip
\texttt{Question:} \qavariable{question}\par
\texttt{Answer format:} \qavariable{answer\_format}\par\medskip
\textbf{Image}\par\smallskip
\qavariable{chart\_image}\par\medskip
\textbf{Additional input for scene-graph-augmented prompting}\par\smallskip
\qavariable{scene\_graph\_json}
\end{qapromptbox}
The answer-format instruction specifies the required answer form, units, precision, or list order and is shared across prompting methods.
For guided prompting, it applies to \texttt{final\_answer}.
Box titles and placeholders are presentation aids, not additional model instructions.

\suppressfloats[t]
\begin{qapromptbox}[label=box:qa-guided,breakable=false]{Shared system instruction: Guided methods}
Answer the chart question using the image and any supplied scene graph. Follow the requested answer format exactly and preserve necessary units. Use boxes, group membership, and spatial relations to associate the relevant entities, marks, and values; use node text when available and otherwise read it from the image. Node identifiers, coordinates, and distances are not measured chart values. Bounding boxes use integer [x\_min,y\_min,x\_max,y\_max] in original-image pixels, with the origin at the top left. Directions describe the source box relative to the target box; alignment and overlap describe geometry, not semantic identity. Use only the relevant evidence and trust the image if annotations conflict with it. 
\par\smallskip
Locate the evidence relevant to the question in the image first. Then use the supplied annotations, when available, to associate labels, data marks and values within the appropriate groups. Record those key elements and groups before giving the result of the requested lookup, comparison or calculation. When annotations contain no text, read the text and values from the image.
\par\smallskip
For this condition, use the JSON output format below instead of a final-answer-only response. Apply the question's answer-format instruction, including units, precision and list order, only inside the final\_answer string. Return exactly one JSON object, with no Markdown or surrounding prose. Do not include extended reasoning.
First output a compact evidence record, then the final\_answer field. Use this form:

\begin{lstlisting}[style=qajson]
{
  "evidence": [{
    "entity": "relevant chart entity or mark",
    "group": "its relevant visual group or series",
    "node_refs": ["exact supplied node labels, when available"],
    "observed_value": "visible evidence, with units when relevant"
  }],
  "final_answer": "the requested answer"
}
\end{lstlisting}

Keep the evidence list focused on the key elements needed for this question. It is an evidence record, not a detailed reasoning transcript. Use only node references actually present in the supplied annotations; if no annotations are supplied, use an empty node\_refs list and identify the evidence and groups from the image. Do not copy coordinates or spatial distances as measured chart values.
\end{qapromptbox}
\section{Scene-Graph-Conditioned Image Generation}
\label{app:generation}

\subsection{Model-Specific Input Construction}
\label{app:generation-inputs}
All models generate from textual scene graph specifications.
Reference-image pixels are not supplied as conditioning.
The scoring-stage hierarchy reconstruction described below can change group counts.
The evaluation always compares the generated result with the reference graph.

\paragraph{Native JSON inputs: Ideogram 4 and Qwen-Image-SP.}
We convert the scene graph to each model's native JSON format, without an additional natural-language instruction prefix.
Ideogram 4 receives all source nodes as caption entries, with \texttt{type=text} for text and \texttt{type=obj} for other nodes.
It preserves literal text and normalized $[y_0,x_0,y_1,x_1]$ boxes.
Colors beyond the native palette capacity and RGBA opacity are retained in descriptive text.
For Qwen-Image-SP, element boxes become \texttt{position} strings in \texttt{<bbox>left top right bottom</bbox>} format; literal text and visual descriptions populate the \texttt{caption} and \texttt{appearance} fields.
Identifiers in descriptive prose are adapted to the retained records without changing visible text.

\paragraph{Free-form inputs: all other models.}
The remaining models receive rendering instructions followed by a serialized scene graph, preserving literal text, normalized bounding boxes, and the distinction between layout groups and visible elements.
For each model, the input token budget follows either the hard limit of its inference interface or the largest budget found to support stable image generation in preliminary experiments.
These limits characterize the model's practical capacity to process structured specifications under the evaluated configuration.
When the input exceeds this budget, we truncate the serialized scene graph to a breadth-first prefix of complete nodes that fits within the limit.
Evaluation remains against the full reference scene graph, so the reported scores reflect end-to-end performance, including both input-capacity limitations and the preservation of supplied content.
The recorded free-form inputs use $[x_0,y_0,x_1,y_1]$ boxes, converted from the paper's $[y_0,x_0,y_1,x_1]$ convention.

\subsection{Generation Configurations}
\label{app:generation-configurations}
\paragraph{Open-source generation settings.}
Table~\ref{tab:app-generation-open-settings} lists the checkpoints and sampling settings.
The seed is $42+i$, where $i$ is the sample's zero-based index in the fixed 1,000-image manifest.
Generation uses one output per model and sample, with no selection among multiple candidate images.
The six non-Ideogram local models use bfloat16.

\begin{table}[!htbp]
\centering
\caption{Recorded checkpoints and generation settings. CFG denotes classifier-free guidance.}
\label{tab:app-generation-open-settings}
\small
\renewcommand{\arraystretch}{1.15}
\begin{tabular}{@{}>{\raggedright\arraybackslash}p{0.23\linewidth}>{\raggedright\arraybackslash}p{0.48\linewidth}>{\raggedright\arraybackslash}p{0.23\linewidth}@{}}
\toprule
Model & Checkpoint & Sampling \\
\midrule
HiDream-O1-Image & \path{HiDream-ai/HiDream-O1-Image} & 50 steps; CFG 5; shift 3; UniPC \\
GLM-Image & \path{zai-org/GLM-Image} & 50 steps; CFG 1.5 \\
Qwen-Image-2512 & \path{Qwen/Qwen-Image-2512} & 50 steps; CFG 4 \\
Qwen-Image-SP & \path{Qwen/Qwen-Image} with \path{heheyas/Qwen-Image-SP} transformer and text-encoder overlays & 25 steps; CFG 4 \\
ERNIE-Image-Turbo & \path{baidu/ERNIE-Image-Turbo} & 8 steps; CFG 1; prompt enhancement disabled \\
SenseNova-U1 Infographic V3 & \path{sensenova/SenseNova-U1-8B-MoT-Infographic-V3} & 50 steps; CFG 4; shift 3; thinking disabled \\
Ideogram 4 & Ideogram 4, recorded ComfyUI Quality48 workflow & 48 Euler steps; guidance 7 for 45 steps, then 3 for 3 steps; $\mu=0$, $\sigma=1.5$ \\
\bottomrule
\end{tabular}
\end{table}

Qwen-Image-2512 uses a fixed negative prompt discouraging poor rendering quality, malformed text, and distorted chart structure.
Qwen-Image-SP uses an empty negative prompt.
SenseNova uses CFG over the full sampling interval with no CFG normalization.
GLM's glyph input contains the literal text of retained nodes; its semantic input contains the compact JSON.
The recorded GLM implementation samples the required image-code grids before emitting the terminal end-of-sequence token, and treats glyph-capacity overflow as a failure instead of truncating text.

\paragraph{Proprietary models.}
The GPT-Image API identifiers are \texttt{gpt-image-2} and \texttt{gpt-image-2.5-sunburst}.
Requests use \texttt{size=auto}, \texttt{quality=high}, and PNG output.
The Nano Banana identifiers are \texttt{gemini-3.1-flash-image} and \texttt{gemini-3-pro-image} for Nano Banana 2 and Pro, respectively.
Both use Vertex \texttt{generate\_content} with image-only response modality.
Image size, maximum output tokens, temperature, and candidate count are left at provider defaults in these requests.
The proprietary requests do not set a random seed.

\paragraph{Graph extraction and scoring.}
Generated images are annotated using the shared procedure in Appendix~\ref{app:annotation}; the reference graph is not supplied to this extraction stage.
Element text, boxes, types, and colors remain fixed during this grouping step.
Matching and attribute similarities follow Appendix~\ref{app:i2sg}.
Overall Node, Hierarchical, and Spatial scores are averaged across the 1,000 images; individual attribute columns pool applicable matched-node scores.

\subsection{Further Analysis}
\label{app:generation-analysis}
\paragraph{Generation performance by input graph size.}
We analyze how generation performance varies with the number of nodes in the input scene graph, counting elements and groups and excluding \texttt{ROOT}.
Node counts are measured before model-specific input construction.
Using the same bin boundaries as the image-to-scene-graph analysis, we divide the 1,000 inputs into $n<50$, $50\leq n<75$, $75\leq n<100$, $100\leq n\leq125$, and $n>125$, containing 191, 218, 193, 178, and 220 images, respectively.
All models use the same images within each bin, including failed outputs, and all three scores are averaged equally over images.
Spatial uses the same Boolean proximity definition as the main results.

\begin{figure}[!htbp]
\centering
\includegraphics[width=\linewidth]{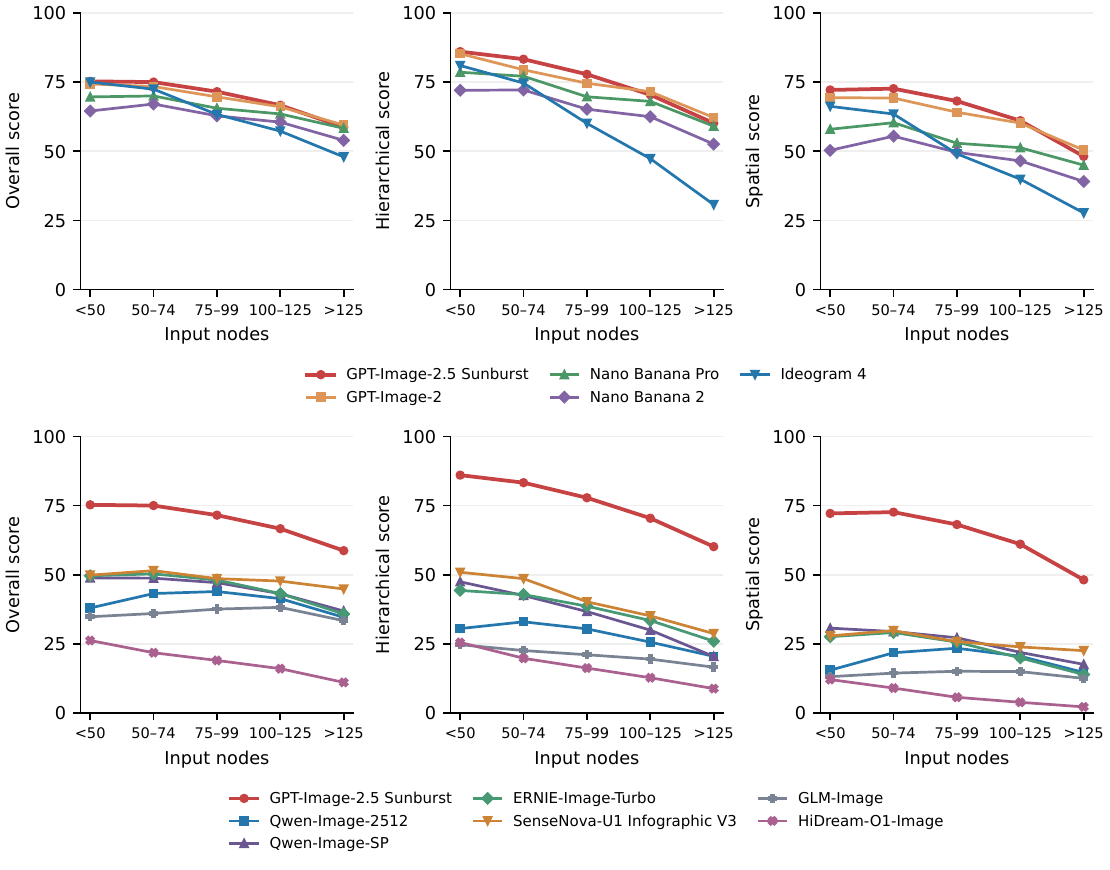}
\caption{Generation performance by input node count.
The top row compares the GPT-Image and Nano Banana variants with Ideogram 4; the bottom row includes the remaining six models, with GPT-Image-2.5 Sunburst repeated for reference.
Each point is the mean score over the images in its bin.}
\label{fig:generation-complexity-supp}
\end{figure}

Figure~\ref{fig:generation-complexity-supp} reports all eleven models.
Every model has lower Overall, Hierarchical, and Spatial scores in the largest-node bin than in the smallest-node bin, although the intermediate curves are not always monotonic.
GPT-Image-2.5 Sunburst and Ideogram 4 have similar Overall scores for $n<50$ (75.3 versus 75.0), while their gap increases to 10.7 points for $n>125$ (58.7 versus 48.0).
Their Hierarchical gap increases from 5.0 to 29.5 points between these two bins, and their Spatial gap increases from 5.9 to 20.4 points.
Both models lose accuracy on larger graphs, but Sunburst retains a greater advantage over Ideogram 4 in preserving elements and relationships.
Because image-source composition also varies across bins, these curves characterize performance across graph sizes rather than isolate a causal effect of node count.

\paragraph{Small boxes and positional sensitivity.}
In the final evaluation references, 71,721 of 99,010 non-root nodes (72.4\%) have boxes occupying at most 0.5\% of the image area.
Across the eleven reported models, these nodes account for 501,672 of 786,305 matched pairs (63.8\%).
Their mean normalized GIoU similarity is 24.9, compared with 54.8 for larger boxes, on the 0--100 reporting scale.
These are conditional means over matched pairs, rather than detection recalls.
Small regions have lower Box scores and are sensitive to displacements that remove much of their overlap.
Box should therefore be considered alongside text and relationship preservation when assessing a generated chart.

\paragraph{Spatial components.}
Table~\ref{tab:app-generation-spatial} separates the four spatial components using the final matching and image-level aggregation.
Alignment is the lowest-scoring component for every model.
For example, GPT-Image-2.5 Sunburst scores 68.5 for direction but 43.0 for alignment; Ideogram 4 scores 53.0 and 30.2, respectively.
Preserving broad directional organization therefore does not imply preserving exact shared edges or centers.
Proximity measures positive Boolean near-pair F1, with no credit for pairs that are non-near on both sides, using the geometric rule in Appendix~\ref{app:spatial-relations}.

\begin{table}[!htbp]
\centering
\caption{Spatial relationship scores by component, averaged over all 1,000 images per model. Each column is multiplied by 100. The overall Spatial score in Table~\ref{tab:sg_preservation} pools relationship counts within each image; it is not the arithmetic mean of these four columns.}
\label{tab:app-generation-spatial}
\small
\begin{tabular}{@{}lrrrr@{}}
\toprule
Model & Direction & Alignment & Overlap & Proximity \\
\midrule
HiDream-O1-Image & 6.7 & 3.5 & 13.5 & 5.0 \\
GLM-Image & 16.0 & 3.6 & 19.7 & 7.5 \\
Qwen-Image-2512 & 21.9 & 5.2 & 25.8 & 11.1 \\
Qwen-Image-SP & 28.2 & 9.1 & 34.1 & 15.2 \\
ERNIE-Image-Turbo & 25.5 & 9.5 & 34.0 & 15.2 \\
SenseNova-U1 Infographic V3 & 28.2 & 11.1 & 36.3 & 16.8 \\
Ideogram 4 & 53.0 & 30.2 & 55.1 & 36.8 \\
Nano Banana 2 & 51.5 & 28.8 & 57.9 & 35.3 \\
Nano Banana Pro & 57.3 & 32.6 & 62.7 & 39.0 \\
GPT-Image-2 & 67.2 & 39.3 & 69.0 & 45.4 \\
GPT-Image-2.5 Sunburst & 68.5 & 43.0 & 69.7 & 49.7 \\
\bottomrule
\end{tabular}
\end{table}

\end{document}